\documentclass{article}
\usepackage{ijcai25}
\usepackage{times}
\usepackage{soul}
\usepackage{url}
\usepackage{natbib}
\usepackage[hidelinks]{hyperref}
\usepackage[utf8]{inputenc}
\usepackage[small]{caption}
\usepackage{graphicx}
\usepackage{amsmath}
\usepackage{amsthm}
\usepackage{booktabs}
\usepackage[switch]{lineno}

\usepackage{algorithm}
\usepackage[compatible]{algpseudocode}  

\usepackage{listings}
\title{Sketch Decompositions for Classical Planning via Deep Reinforcement Learning}

\newif\ifanonymous
\anonymousfalse

\newif\ifappendix
\appendixfalse

\ifanonymous
    \author{
    Anonymous Submission
    }
\else
    \author{
    Michael Aichm\"uller
    \and
    Hector Geffner
    \affiliations
    RWTH Aachen University, Germany
    \emails
    michael.aichmueller@ml.rwth-aachen.de,
    hector.geffner@ml.rwth-aachen.de
    }
\fi

\usepackage{thmtools}
\usepackage{thm-restate}
\usepackage{booktabs}
\usepackage{enumerate}
\usepackage{xspace}
\usepackage{todonotes}
\usepackage{amssymb}
\usepackage{amsmath}
\usepackage{mathtools}
\usepackage{multirow}
\usepackage[title]{appendix}
\presetkeys{todonotes}{inline}{}
\newcommand{\michael}[1]{}
\newcommand{\hector}[1]{}

\newcommand{\inlinecite}[1]{\citet{#1}}

\newcommand{\tup}[1]{(#1)}

\newcommand{\iw}[1]{\ensuremath{\text{IW}(#1)}\xspace}
\newcommand{\iwraw}{\ensuremath{\text{IW}}\xspace}

\newcommand{\siw}{\ensuremath{\text{SIW}}\xspace}

\newcommand{\siwk}{\ensuremath{\text{SIW}^\pi(k)}\xspace}
\newcommand{\siwone}{\ensuremath{\text{SIW}^\pi(1)\xspace}}
\newcommand{\siwtwo}{\ensuremath{\text{SIW}^\pi(2)\xspace}}

\newcommand{\Q}{\mathcal{Q}}

\newcommand{\Omit}[1]{}

\newcommand{\domain}{\ensuremath{D}}
\newcommand{\instance}{\ensuremath{I}}
\newcommand{\goal}{\ensuremath{\mathit{Goal}}}
\newcommand{\initial}{\ensuremath{\mathit{Init}}}

\newcommand{\states}{\ensuremath{S}}
\newcommand{\initialstate}{\ensuremath{s_0}}
\newcommand{\goalstates}{\ensuremath{G}}
\newcommand{\actions}{\ensuremath{\mathit{Act}}}
\newcommand{\successor}{\ensuremath{f}}
\newcommand{\applicability}{\ensuremath{A}}

\newcommand{\sat}[2]{\ensuremath{\mathit{at}(#1,#2)}}

\newcommand{\gat}[2]{\ensuremath{\mathit{at}(#1, #2)}}

\newcommand{\bon}[2]{\ensuremath{\mathit{on}(#1, #2)}}

\newcommand*{\eg}{e.g.\@\xspace}
\newcommand*{\ie}{i.e.\@\xspace}

\newlength{\firstword}

\newlength{\firstwordagain}
\newcommand{\stretchyr}[2]{%
  \settowidth{\firstwordagain}{#1}%
  \makebox[\firstwordagain][r]{#2}%
}

\DeclareMathOperator*{\argmax}{arg\,max}

\newcounter{listing}[section]
\renewcommand{\thelisting}{\arabic{listing}}

\newcommand{\listingcaption}[1]{%
    \refstepcounter{listing}%
    \centering \normalsize \textnormal{Listing \thelisting: #1}\par}

\begin{document}

\maketitle

\ifappendix
    \appendix
\section{Appendix}
This section provides a detailed description of the evaluation data, including the domains and key characteristics of the instances. We also describe additional predicates that were necessary for successful training. Furthermore, we present results for greedily chosen decompositions, additional baseline details, and an experiment assessing the generalization capability of \siwone\ in the Delivery domain. Finally, we highlight generated plans from each domain that corroborate our analysis of the sketch decompositions learned by our method.

\subsection{Data Details}
We explicate in the following in greater detail our evaluation domains and provide richer information about the characteristics of our training and testing instances. The number of training instances used are mentioned in parentheses. In general, while our data is based on prior work \citep{simon:kr2023,drexler:icaps2022},  we extended this data to ensure a uniform number of $40$ testing instances.
\begin{itemize}
    \item \textbf{Blocks (19, 22)}: This domain includes two goal variations. The first involves permuting an already constructed tower (19), while the second requires building multiple towers (22) from a starting position of several small towers and free blocks. The goals include only $\mathit{on}$ atoms. The training, validation, and test sets use instances from the IPC, featuring $4$-$7$ blocks for training, $7$ blocks for validation, and $8$-$26$ blocks for testing in single tower, and $8$-$27$ in multiple tower problems. In the latter case we also have problems with $2$-$3$ towers in the training set, while the test set varies between $2$ to $5$ target towers.
    
    \item \textbf{Childsnack (49)}: This domain focuses on planning how to make and serve sandwiches to a group of children, some of whom are allergic to gluten. There are two actions for making sandwiches: one that assembles a regular sandwich and another that ensures all ingredients are gluten-free. Additional actions include placing sandwiches on trays, moving trays, and serving the sandwiches. Problems in this domain define the initial ingredients available for making sandwiches, and the goal is to ensure that every child is served a sandwich they can safely eat, considering their allergies. The instances vary in the number of children, the ratio of gluten-allergic children, and the contents of sandwiches. The 49 training instances involve $1$-$3$ children with a randomly allotted gluten-ratio and $1$-$4$ trays, while test instances go up to $4$ children and trays.
   
    \item \textbf{Delivery (75)}: This problem revolves around picking up objects in a grid and delivering them one by one to a target cell, with no obstacles present. Training Instances are defined on grids up to $9 \times 9$ in size, containing up to $4$ packages, a single agent, and a shared target cell. The training set is generally made up of smaller grids and single package problems for larger grids, while test sets feature combinations of grids up to size $10 \times 10$ and up to $10$ packages with unique destinations for each package.
    
    \item \textbf{Grid (13)}: The objective is to find keys, unlock doors, and place specific keys at designated locations. Dropoff locations for some keys may, however, be blocked by locked doors. Training problems vary in grid sizes with $x \times y \in \{1 \times z \mid z = 3,5,7,9\} \cup \{3 \times z\mid z = 5,7,9\} \cup \{ 5 \times 7 \}$, with up to $3$ locks and lock shapes, and $4$ keys. For testing, the complexity increases with grid sizes up to $9 \times 9$, featuring up to $9$ keys, $10$ locked locations, and $5$ different lock shapes.
    
    \item \textbf{Gripper (10)}: The task requires moving balls from one room to another using a robot equipped with two grippers. Training instances feature $1$-$9$ balls, validation instances have 10 balls, and test instances include $22$-$61$ balls.
    
    \item \textbf{Logistics (25)}: In this domain, the task involves moving packages between cities of different sizes using trucks and airplanes. Instances vary between $1$-$2$ airplanes, $1$-$5$ cities with $2$ locations each, and $1$-$6$ packages, and $1$-$5$ trucks. For testing we reach up to $3$ airplanes, $9$ packages, $5$ trucks, and $5$ cities with $3$ locations. 
    
    \item \textbf{Miconic (63)}: Elevator operations to pick up and deliver passengers across different floors have to be scheduled. Training and validation instances vary between $2$-$8$ floors and $1$-$5$ passengers. The test set extends up to $60$ floors and $30$ passengers.
    
    \item \textbf{Reward (15)}: The goal is to navigate a grid with obstacles to collect all rewards. Instances are defined on square grids with widths ranging from $3$-$7$ for training and validation, and $10$-$25$ for testing. The maximum number of rewards varies from $6$ for training and validation to $23$ for testing.
    
    \item \textbf{Spanner (140)}: "Bob" has to tighten nuts at one end of a one-way corridor by collecting spanners along the way. The number of locations varies between $1$-$20$ across training and validation with up to $6$ spanners, $4$ nuts, and a maximum surplus of $3$ spanners in any instance. For testing, we vary instances with up to and $12$ nuts, $24$ spanners, $20$ locations, and a maximum surplus of $12$ spanners. 
    
    \item \textbf{Visitall (12)}: The objective is to visit all cells in a grid with no obstacles. The training grid sizes are limited to $3 \times 3$ or $4 \times 2$, while the validation set uses $5 \times 2$ grids. The test set includes larger grids ranging from $4 \times 4$ to $10 \times 10$ with a uniform randomly drawn fraction of all cells to visit.
\end{itemize}

\subsection{Derived Predicates}
Enlarging the neighborhood definition from $N(s)$ to $N_k(s)$ initially prevented us from deriving decompositions, likely due to the limited expressivity of the relational GNN architecture \citep{drexler:kr2024}. Our approach represents a state s by the atoms $p(x_1, \ldots, x_{a(p)})$ true in $s$, where $a(p)$ is the predicate arity, and by goal atoms $p_g$ for the conjunctive goal. To increase expressivity, we introduced domain-independent derived predicates $p_{\mathrm{ug}} \coloneqq p_g \wedge \neg p$ for each goal predicate $p_g$, indicating unsatisfied goals.

\subsection{Greedy Decomposition Results}

We present in table \ref{tbl:experiments_greedy} the results of \siwk with $k = 1, 2$ and subgoal states $G_k^\pi(s)$ chosen \textit{greedily} according to the learned general policy $\pi( \cdot \mid s)$, \ie 
\begin{equation}
    G_k^{\pi}(s) \coloneqq   \{ \, s' \,\} \, , \, s' = \argmax_{s' \in N_k(s)} \pi(s' \mid s). \notag
\end{equation}
Ties are broken randomly.
The table is organized again with no subgoal cycle prevention to the left and subgoal cycle prevention to the right half. The top segment shows \siwone\ and the bottom \siwtwo. Overall, there are no significant differences to stochastically sampled subgoals: One fewer instance is solved in Blocksworld Multiple (width-1, no cycle prevention), while one more in Logistics (width-1, with cycle prevention). Plan-lengths are at most changed by a couple of actions on average. We can conclude that the model selected subgoals with high confidence. Indeed, observing planning executions with confidence outputs showcases most subgoals are sampled with 100\% probability. Ambiguity is observed mostly for equivalent subgoal choices, such as which package to pickup next for width-1 decompositions in Delivery, and for subgoals in domains whose decompositions are not readily interpretable (Blocksworld/Logistics).
\begin{table*}[ht!]
  \centering
  \footnotesize
  \setlength{\tabcolsep}{2.25pt}

\begin{tabular}{@{\extracolsep{4.8pt}}lrrrrlrrrlr}
\toprule
& \multicolumn{1}{c}{LAMA} & \multicolumn{4}{c}{No Cycle Prevention} & \multicolumn{4}{c}{Subgoal Cycle Prevention} & \multicolumn{1}{c}{Validation} \\ \cmidrule{2-2}\cmidrule{3-6} \cmidrule{7-10} \cmidrule{11-11}
Domain (\#) & Cov. (\%) $\uparrow$ & Cov. (\%) $\uparrow$ & SL $\downarrow$ & L $\downarrow$ & PQ = L / L$_M$  $\downarrow$ & Cov. (\%) $\uparrow$ & SL $\downarrow$& L $\downarrow$ & PQ = L / L$_M \downarrow$ & L$_V$ / L$_V^* \downarrow$ \\
\midrule
&&&&&&& \\
    [-.5em]
& \multicolumn{9}{c}{\textbf{\siwone}} &\\
\midrule
\midrule
Blocks (40) & 40 (\stretchyr{100}{100} \%) & 39 (\stretchyr{100}{98} \%) & 21 & 80 & 1.05 = 80 / 76 & 40 (\stretchyr{100}{100} \%) & 21 & 81 & ~~1.05 = 81 / 77 & 1.22\\
Blocks-mult. (40) & 39 (\stretchyr{100}{98} \%) & 31 (\stretchyr{100}{78} \%) & 18 & 55 & 1.06 = 55 / 52 & 39 (\stretchyr{100}{98} \%) & 23 & 68 & ~~1.15 = 66 / 57 & 1.32\\
Childsnack (40) & 40 (\stretchyr{100}{100} \%) & 40 (\stretchyr{100}{100} \%) & 6 & 11 & 1.06 = 11 / 10 & 40 (\stretchyr{100}{100} \%) & 6 & 11 & ~~1.06 = 11 / 10 & 1.00\\
Delivery (40) & 40 (\stretchyr{100}{100} \%) & 40 (\stretchyr{100}{100} \%) & 10 & 52 & 1.02 = 52 / 50 & 40 (\stretchyr{100}{100} \%) & 10 & 52 & ~~1.02 = 52 / 50 & 1.00\\
Grid (40) & 38 (\stretchyr{100}{95} \%) & 23 (\stretchyr{100}{58} \%) & 7 & 39 & 1.19 = 39 / 33 & 38 (\stretchyr{100}{95} \%) & 71 & 359 & 10.21 = 359 / 35 & 11.85\\
Gripper (40) & 40 (\stretchyr{100}{100} \%) & 40 (\stretchyr{100}{100} \%) & 83 & 164 & 1.32 = 164 / 124 & 40 (\stretchyr{100}{100} \%) & 83 & 164 & ~~1.32 = 164 / 124 & 1.00\\
Logistics (40) & 38 (\stretchyr{100}{95} \%) & 10 (\stretchyr{100}{25} \%) & 8 & 19 & 1.30 = 16 / 12 & 25 (\stretchyr{100}{62} \%) & 133 & 223 & 12.93 = 235 / 18 & 60.36\\
Miconic (40) & 40 (\stretchyr{100}{100} \%) & 40 (\stretchyr{100}{100} \%) & 32 & 61 & 1.16 = 61 / 52 & 40 (\stretchyr{100}{100} \%) & 32 & 61 & ~~1.16 = 61 / 52 & 1.00\\
Reward (40) & 40 (\stretchyr{100}{100} \%) & 40 (\stretchyr{100}{100} \%) & 15 & 195 & 2.30 = 195 / 85 & 40 (\stretchyr{100}{100} \%) & 15 & 195 & ~~2.30 = 195 / 85 & 1.00\\
Spanner (40) & 30 (\stretchyr{100}{75} \%) & 40 (\stretchyr{100}{100} \%) & 24 & 43 & 1.29 = 40 / 31 & 40 (\stretchyr{100}{100} \%) & 24 & 43 & ~~1.29 = 41 / 31 & 1.00\\
Visitall (40) & 40 (\stretchyr{100}{100} \%) & 40 (\stretchyr{100}{100} \%) & 8 & 68 & 1.65 = 68 / 41 & 40 (\stretchyr{100}{100} \%) & 8 & 68 & ~~1.65 = 68 / 41 & 1.03\\
&&&&&&&&&&\\[-.5em]
& \multicolumn{9}{c}{\textbf{\siwtwo}} &\\
\midrule
\midrule
Blocks (40) & 40 (\stretchyr{100}{100} \%) & 40 (\stretchyr{100}{100} \%) & 9 & 134 & 1.73 = 134 / 77 & 40 (\stretchyr{100}{100} \%) & 9 & 134 & 1.73 = 134 / 77 & 1.27\\
Blocks-mult. (40) & 39 (\stretchyr{100}{98} \%) & 40 (\stretchyr{100}{100} \%) & 8 & 78 & 1.35 = 78 / 58 & 40 (\stretchyr{100}{100} \%) & 8 & 78 & 1.35 = 78 / 58 & 1.07\\
Childsnack (40) & 40 (\stretchyr{100}{100} \%) & 40 (\stretchyr{100}{100} \%) & 3 & 13 & 1.21 = 13 / 10 & 40 (\stretchyr{100}{100} \%) & 3 & 13 & 1.21 = 13 / 10 & 1.00\\
Delivery (40) & 40 (\stretchyr{100}{100} \%) & 40 (\stretchyr{100}{100} \%) & 5 & 57 & 1.12 = 57 / 50 & 40 (\stretchyr{100}{100} \%) & 5 & 57 & 1.12 = 57 / 50 & 1.00\\
Grid (40) & 38 (\stretchyr{100}{95} \%) & 38 (\stretchyr{100}{95} \%) & 3 & 47 & 1.34 = 47 / 35 & 38 (\stretchyr{100}{95} \%) & 3 & 47 & 1.34 = 47 / 35 & 1.00\\
Logistics (40) & 38 (\stretchyr{100}{95} \%) & 40 (\stretchyr{100}{100} \%) & 4 & 34 & 1.33 = 34 / 26 & 40 (\stretchyr{100}{100} \%) & 4 & 34 & 1.33 = 34 / 26 & 1.01\\
    \bottomrule
\end{tabular}
\caption{
\small   Performance metrics for  learned  general decompositions  per domain and width with \emph{greedy decompositions} for comparison with stochastic decompositions (main paper body).
  Rows show results over 40 test instances per domain. Coverage (Cov.) indicates the number of successful plans (fraction in parentheses). SL and L represent
  average subgoal and plan lengths  (rounded to nearest integer), while L$^*$ and L$_M$ represent optimal and LAMA plan lengths resp. averaged. 
  Plan quality (PQ) is the ratio of average plan lengths  to those of LAMA.
  Validation score L$_V$ / L$_V^*$ is shown on the right. 
  Arrows indicate preferred metric directions. The only difference to stochastic decomposition results is that one fewer instance is solved in Blocksworld Multiple (width-1, no cycle prevention), while one more in Logistics (width-1, with cycle prevention). Plan-lengths differences are insignificant.
}
  \label{tbl:experiments_greedy}
\end{table*}

\subsection{Baseline Comparison Details}
For our baseline comparison, we retrained the method of \citet{simon:kr2023} (on which our work is founded) to produce width-0 decompositions. We replicated their setup as closely as possible, extending training from 12 to 24 hours on smaller A10 GPUs, and added unsatisfied-goal predicates $p_{\mathrm{ug}}$ to match our approach. Among five training runs, we selected the best model via validation performance. Table~\ref{tbl:iw0_baseline} shows this baseline fails to generalize in Grid, Logistics (as expected), and Reward - this, likely because a grid size of up to $25 \times 25$ in Reward exceeds the GNN’s reasoning limit imposed by the message-passing layer count $L=30$. By contrast, our method circumvents this through \(\iw{k}\) searches, extending its effective reasoning depth.

In general, \citet{simon:kr2023} exhibits less consistency: about a quarter of Visitall and a tenth of Delivery, Gripper, and Spanner tests fail with subgoal cycle prevention (and failures rise further without it). In contrast, \siwk\ is more robust, though the baseline’s direct plan-cost optimization yields shorter plans (Blocks, Gripper, Reward, Visitall). Surprisingly, it marginally outperforms LAMA in Delivery and significantly in Blocksworld.

\begin{table*}[h!]
  \centering
  \footnotesize
  \setlength{\tabcolsep}{2.25pt}

\begin{tabular}{@{\extracolsep{4.8pt}}lrrrrlrrrlr}
\toprule
& \multicolumn{1}{c}{LAMA} & \multicolumn{4}{c}{No Cycle Prevention} & \multicolumn{4}{c}{Subgoal Cycle Prevention} & \multicolumn{1}{c}{Validation} \\ \cmidrule{2-2}\cmidrule{3-6} \cmidrule{7-10} \cmidrule{11-11}
Domain (\#) & Cov. (\%) $\uparrow$ & Cov. (\%) $\uparrow$ & SL $\downarrow$ & L $\downarrow$ & PQ = L / L$_M$  $\downarrow$ & Cov. (\%) $\uparrow$ & SL $\downarrow$& L $\downarrow$ & PQ = L / L$_M \downarrow$ & L$_V$ / L$_V^* \downarrow$ \\
&&&&&&&&&&\\[-.5em]
\midrule
\midrule
Blocks (40) & 40 (\stretchyr{100}{100} \%) & 40 (\stretchyr{100}{100} \%) & 54 & 54 & 0.70 = 54 / 77 & 40 (\stretchyr{100}{100} \%) & 54 & 54 & ~~0.70 = 54 / 77 & 1.00\\
Blocks-mult. (40) & 39 (\stretchyr{100}{98} \%) & 40 (\stretchyr{100}{100} \%) & 44 & 44 & 0.75 = 43 / 58 & 40 (\stretchyr{100}{100} \%) & 44 & 44 & ~~0.75 = 43 / 58 & 1.00\\
Childsnack (40) & 40 (\stretchyr{100}{100} \%) & 40 (\stretchyr{100}{100} \%) & 11 & 11 & 1.10 = 11 / 10 & 40 (\stretchyr{100}{100} \%) & 11 & 11 & ~~1.10 = 11 / 10 & 1.00\\
Delivery (40) & 40 (\stretchyr{100}{100} \%) & 35 (\stretchyr{100}{88} \%) & 48 & 48 & 0.96 = 48 / 50 & 36 (\stretchyr{100}{90} \%) & 48 & 48 & ~~0.97 = 48 / 50 & 1.04\\
Grid (40) & 38 (\stretchyr{100}{95} \%) & 11 (\stretchyr{100}{28} \%) & 14 & 14 & 1.00 = 14 / 14 & 23 (\stretchyr{100}{58} \%) & 41 & 41 & ~~1.61 = 41 / 25 & 1.00\\
Gripper (40) & 40 (\stretchyr{100}{100} \%) & 38 (\stretchyr{100}{95} \%) & 312 & 312 & 2.58 = 312 / 121 & 36 (\stretchyr{100}{90} \%) & 122 & 122 & ~~1.03 = 122 / 118 & 1.00\\
Logistics (40) & 38 (\stretchyr{100}{95} \%) & 6 (\stretchyr{100}{15} \%) & 12 & 12 & 1.03 = 12 / 11 & 20 (\stretchyr{100}{50} \%) & 607 & 607 & 35.28 = 630 / 18 & 1.00\\
Miconic (40) & 40 (\stretchyr{100}{100} \%) & 40 (\stretchyr{100}{100} \%) & 54 & 54 & 1.04 = 54 / 52 & 40 (\stretchyr{100}{100} \%) & 54 & 54 & ~~1.04 = 54 / 52 & 1.00\\
Reward (40) & 40 (\stretchyr{100}{100} \%) & 8 (\stretchyr{100}{20} \%) & 41 & 41 & 1.01 = 41 / 40 & 20 (\stretchyr{100}{50} \%) & 74 & 74 & ~~1.15 = 74 / 65 & 1.00\\
Spanner (40) & 30 (\stretchyr{100}{75} \%) & 37 (\stretchyr{100}{92} \%) & 38 & 38 & 1.14 = 36 / 32 & 37 (\stretchyr{100}{92} \%) & 38 & 38 & ~~1.14 = 36 / 32 & 1.00\\
Visitall (40) & 40 (\stretchyr{100}{100} \%) & 17 (\stretchyr{100}{42} \%) & 27 & 27 & 1.04 = 27 / 27 & 30 (\stretchyr{100}{75} \%) & 43 & 43 & ~~1.27 = 43 / 34 & 1.00\\
\bottomrule
\end{tabular}
\caption{\small
Performance metrics for learned general policies per domain for the baseline method by \citet{simon:kr2023} with \emph{stochastic successor sampling}.
Rows show results over 40 test instances per domain. Coverage (Cov.) indicates the number of successful plans (fraction in parentheses). SL and L represent average subgoal and plan lengths  (rounded to nearest integer), while L$_M$ and L$^*$  represent average LAMA and optimal plan lengths (validation instances only) respectively. Plan quality (PQ) is the ratio of average plan lengths  to those of LAMA. Validation score L$_V$ / L$_V^*$ is shown on the right. Arrows indicate preferred metric directions. Several differences to stochastic width-1/2 decompositions are noticeable: failures in Logistics, Grid, and Reward, and less consistent generalization abilities resulting in lower coverage on average. Plan-lengths are shorter in turn, sometimes even surpassing LAMA.
}
  \label{tbl:iw0_baseline}
\end{table*}

Our second baseline is the combinatorial approach of \citet{drexler:icaps2022}, which learns rule-based sketches from training data using grammar-extracted features (Table~\ref{tbl:dominik}). These features form the basis of a constraint-satisfaction problem that produces a width-k sketch for each problem class. We tested sketches with $k \in \{1,2\}$, but the method’s combinatorial nature led to timeouts under the original setup when learning on our subsets of data. Consequently, we used sketches provided by the authors, who had trained them on a superset of our domains.

In general, sketches from \citet{drexler:icaps2022} do not generalize to unseen goals, so we verified that their training data matched our goal descriptions. This held for most domains, except Blocksworld, which used only atomic goals. As a result, it failed to handle our conjunctive $\mathrm{on}$-atom goals, leading to complete failure in single-tower Blocksworld and near complete failure in multi-tower scenarios. The method also failed in Grid and Logistics for both widths and was unable to find a width-2 sketch for Childsnack. In contrast, \siwtwo\ found width-2 solutions for Childsnack reliably.

Interestingly, the method’s width-2 sketch for Delivery failed in 10\% of tests, despite no apparent structural differences from the successful cases. In all other domains, coverage was perfect. Plan lengths were typically 10–20\% shorter than those of \siwk\, except in Miconic, where they were longer on average.
\begin{table}[t]
  \centering
  \footnotesize
  \setlength{\tabcolsep}{2.pt}

\begin{tabular}{@{\extracolsep{1.pt}}lrrrrrrl}
\toprule
& \multicolumn{2}{c}{LAMA}  & \multicolumn{4}{c}{Combinatorial Learner}  \\ \cmidrule{2-3}\cmidrule{4-8}
Domain (\#) & Cov. (\%) & $L_M$ &  W & Cov. (\%) & L & PQ \\
\midrule
&&&&&&& \\[-1.5em]
\midrule
\multirow{2}{*}{blocks (40)} & \multirow{2}{*}{40 (\stretchyr{100}{100} \%)} & \multirow{2}{*}{78} & 1 & -- & -- & --\\
 &  &  & 2 & -- & -- & --\\
\multirow{2}{*}{blocks-multiple (40)} & \multirow{2}{*}{39 (\stretchyr{100}{98} \%)} & \multirow{2}{*}{59} & 1 & 3 (\stretchyr{100}{8} \%) & 51 & 1.50\\
 &  &  & 2 & 4 (\stretchyr{100}{10} \%) & 36 & 1.26\\
\multirow{2}{*}{childsnack (40)} & \multirow{2}{*}{40 (\stretchyr{100}{100} \%)} & \multirow{2}{*}{11} & 1 & 40 (\stretchyr{100}{100} \%) & 11 & 0.98\\
 &  &  & 2 & -- & -- & --\\
\multirow{2}{*}{delivery (40)} & \multirow{2}{*}{40 (\stretchyr{100}{100} \%)} & \multirow{2}{*}{51} & 1 & 40 (\stretchyr{100}{100} \%) & 50 & 0.96\\
 &  &  & 2 & 36 (\stretchyr{100}{90} \%) & 55 & 1.08\\
\multirow{2}{*}{grid (40)} & \multirow{2}{*}{38 (\stretchyr{100}{95} \%)} & \multirow{2}{*}{36} & 1 & -- & -- & --\\
 &  &  & 2 & -- & -- & --\\
\multirow{2}{*}{gripper (40)} & \multirow{2}{*}{40 (\stretchyr{100}{100} \%)} & \multirow{2}{*}{125} & 1 & 40 (\stretchyr{100}{100} \%) & 124 & 0.99\\
 &  &  & 2 & 40 (\stretchyr{100}{100} \%) & 165 & 1.32\\
\multirow{2}{*}{logistics (40)} & \multirow{2}{*}{38 (\stretchyr{100}{95} \%)} & \multirow{2}{*}{27} & 1 & -- & -- & --\\
 &  &  & 2 & -- & -- & --\\
\multirow{2}{*}{miconic (40)} & \multirow{2}{*}{40 (\stretchyr{100}{100} \%)} & \multirow{2}{*}{53} & 1 & 40 (\stretchyr{100}{100} \%) & 79 & 1.48\\
 &  &  & 2 & 40 (\stretchyr{100}{100} \%) & 60 & 1.12\\
\multirow{2}{*}{reward (40)} & \multirow{2}{*}{40 (\stretchyr{100}{100} \%)} & \multirow{2}{*}{86} & 1 & 40 (\stretchyr{100}{100} \%) & 92 & 1.08\\
 &  &  & 2 & 40 (\stretchyr{100}{100} \%) & 92 & 1.08\\
\multirow{2}{*}{spanner (40)} & \multirow{2}{*}{30 (\stretchyr{100}{75} \%)} & \multirow{2}{*}{32} & 1 & 40 (\stretchyr{100}{100} \%) & 41 & 1.18\\
 &  &  & 2 & 40 (\stretchyr{100}{100} \%) & 41 & 1.18\\
\multirow{2}{*}{visitall (40)} & \multirow{2}{*}{40 (\stretchyr{100}{100} \%)} & \multirow{2}{*}{42} & 1 & 40 (\stretchyr{100}{100} \%) & 53 & 1.27\\
 &  &  & 2 & 40 (\stretchyr{100}{100} \%) & 53 & 1.27\\
    \bottomrule
\end{tabular}
\caption{
\small Performance metrics for learned sketches per domain and width by the baseline method of \citet{drexler:icaps2022}.
  Rows show results over 40 test instances per domain, split into two subrows for each width (W). Columns shared by two adjacent rows are centered vertically. Coverage (Cov.) indicates the number of successful plans (fraction in parentheses). L represents the
  average plan lengths (rounded to nearest integer), while L$_M$ represents averaged LAMA plan lengths. 
  Plan quality (PQ) is the ratio of average plan lengths  to those of LAMA.
}
  \label{tbl:dominik}
\end{table}

\subsection{Generalization Extent}
\label{sec:generalization}
Policies learned by \siwk\ demonstrated strong generalization to problems beyond their training instances in many domains. To investigate this further, we tested the performance boundaries of \siwone\ in the Delivery domain through problems with up to $20 \times 20$ grids, $1$-$4$ agents, and $1$-$25$ packages with distinct goal locations for each package. In particular, we tested square grid sizes from $3^2$ up to $10^2$ with $1$ to $10$ packages. Beyond $10^2$ grid dimensions we increased the grid only in steps of $5$ and varied the package count from $5$ to $25$ in steps of $5$ as well. Figure \ref{fig:delivery_generalization} showcases the results. The model decomposed the problem consistently into $2N$ subproblems $P[s, G_1(s)]$ and $P[s,G_2(s)]$ where $N$ is the total number of packages to deliver in the goal. Here, $G_1(s)$ are those states in which a single package is held by an agent, but the number of undelivered packages is unchanged to the number in $s$, while $G_2(s)$ are states with a lower undelivered package count and, in turn, no package is held anymore. In consequence, the number of agents is irrelevant to the decompositions. This implies that problems with the same number of agents $a$ fall onto the line of $y = 2(x-a) = 2N$ in figure \ref{fig:delivery_generalization}. As such, $\siwone$ always follows a decomposition implied by sketch $R2$. 

Curiously, sketch decompositions akin to $R4$ have never been observed in the Delivery domain, although both $R2$ and $R4$ would lead to the same number of subgoals in Delivery. In words, the model never chooses to first let all available agents pick up an undelivered package each, before dropping off those packages in sequence. 
In general, \textit{for the purpose of width-1 decompositions}, and the given domain representations, problems in Gripper with $N$ balls are equivalent to problems in Delivery with $2$ agents, grid-size $2 \times 1$ and $N$ packages to deliver from cell $0$ to cell $1$. More precisely, the domains are equivalent in the sense of the minimal number of subproblems the top-level goal sketch $\{N > 0\} \rightarrow \{N \downarrow\}$ can be further decomposed into. Here, $N$ is the number of packages and balls to deliver respectively. In both domains, 2 subgoals of width-1 achieve a reduction of $N$\footnote{Indeed, 2 subgoals are also the minimal number of subgoals that decompose the top-level sketch, since both atoms $\gat{ball}{room}$ and $\gat{pkg}{loc}$ are width-2 goals and thus cannot be achieved by a single width-1 subgoal.}. Despite this, we never observe the sketch $R2$ in Gripper. It is currently unclear whether diversifying the training instances of Delivery with multi-agent problems would lead to the same sketch decompositions as in Gripper or whether the underlying representations influence this decision.

\begin{figure*}[t]
    \centering
    \includegraphics[width=0.9\linewidth]{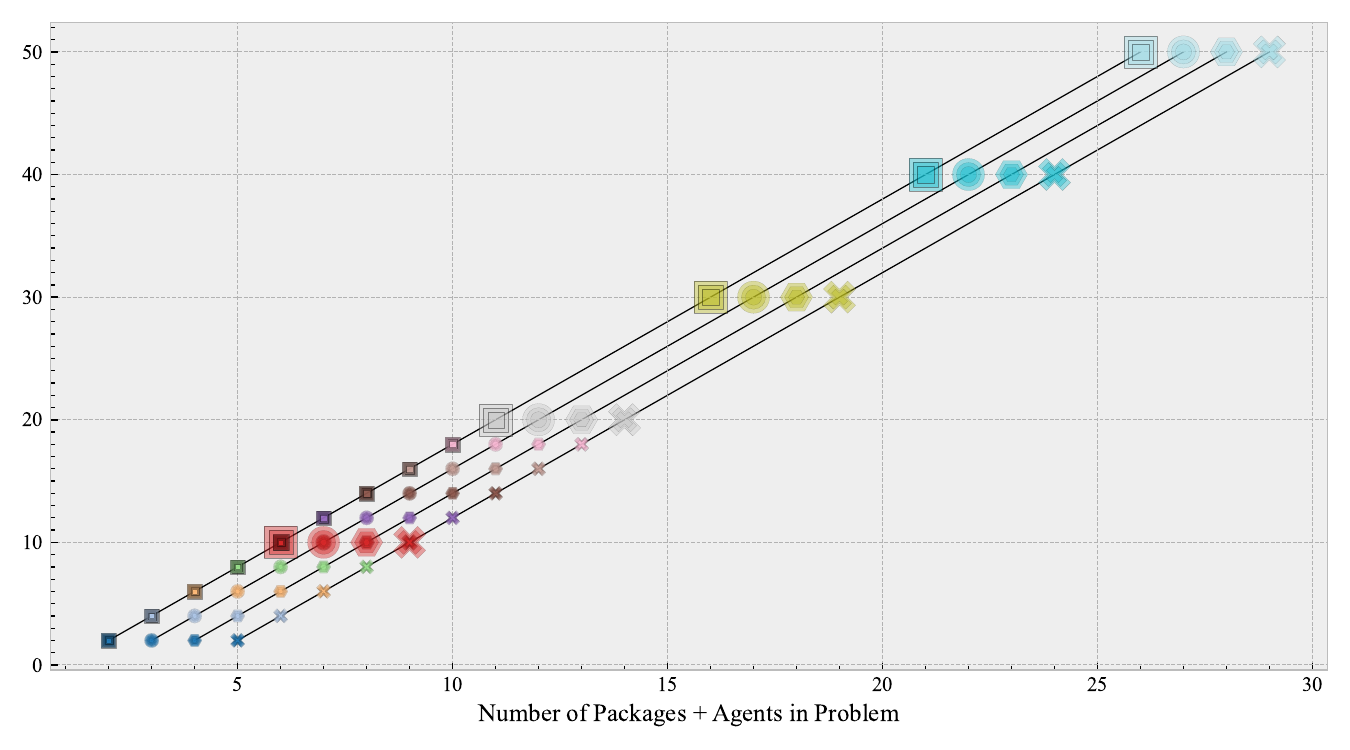}
    \caption{Generalization Performance in the Delivery Domain shown by the number of subgoals selected against the sum of package and agent objects in a problem. Data points represent individual problems with marker sizes proportional to the square grid size. Marker style indicates the number of agents $a=1,2,3,4$ (square, circle, hexagon, cross) involved and marker size is proportional to the cell count in the problem grid. Thus, overlapping points form growing outlines for increasing grid sizes of the problems with the same coordinates. Points of the same color represent the same number of packages in the problem. The lines $y = 2 (x-a)$ correspond to the decompositions for problems with the respective number of agents. Points consistently align with the lines associated with their agent count, indicating that the policy decomposes akin to sketch $R2$ and ignores grid size as well as agent count.}
    \label{fig:delivery_generalization}
\end{figure*}

\subsection{Sketch Analysis From Generated Plans}
We provide example plans generated by \siwone\  at the end of the appendix (see "Structure of \siwk Plans") for the domains we analyzed in the main body. In the following, we use these plans to provide more details on the rationale behind sketch identifications as well as further details on the generated plans in each of the 6 domains. We also add example traces for Blocksworld from our \siwone\ and \siwtwo\ model. Even if approximate sketches could not be identified in this domain, general remarks on the differences between width-1 and width-2 decompositions can be emphasized.

\paragraph{Delivery}
Listings \ref{lst:delivery_2} and \ref{lst:delivery_3} show solutions to delivery problems with 4 and 8 packages. In each problem, first, a subgoal ensures an agent gets hold of a package, represented by the action-chain ending on $\mathit{pick\text{-}package}$, followed by the subgoal delivering said package, represented by the final action $\mathit{drop\text{-}package}$. These two subgoals are iterated, until all packages are delivered. Hence, the model has to learn a boolean feature that captures whether a package is held in a state, while also counting the number of undelivered packages. The sketch $R2$ incorporates these features in a ruleset. 

\paragraph{Gripper}
Listings \ref{lst:gripper_1} and \ref{lst:gripper_3} show plans for Gripper problems with 22 and 55 balls. First, the subgoals which pickup balls in each gripper are selected, represented by paths ending with the action $\mathit{pick}$. Once no gripper is free, the model alternates between dropping off a ball and picking up a new ball. Sketch $R4$ represents these rules. At no point does the model free both grippers indicating that it requires the equality condition for $N_2 = 0$ in the second rule, and not $N_2 \geq 0$ which would allow to empty both grippers before grabbing another undelivered ball. This inefficiency in terms of pickup and dropoff leads to primitive plans consistently longer than optimal due to the extra $\mathit{move}$ actions.

\paragraph{Childsnack}
Childsnack plans (listings \ref{lst:childsnack_1}, \ref{lst:childsnack_2}, \ref{lst:childsnack_3}, \ref{lst:childsnack_4}) can achieve two subgoals which reduce the number of unserved children after each combined execution: The first subgoal which makes a (gluten-free) sandwich and puts it on a tray, and a second which moves and serves the sandwich from the tray to the correct child and table. Sketch $R3$ describes the exhaustion of a first counter, followed by the conditional exhaustion of a second counter. This sketch holds for Childsnack with the first counter being the sandwiches left to make, and the second counter enumerating the unserved children.

\paragraph{Miconic}
In Miconic, the listings \ref{lst:miconic_1}, \ref{lst:miconic_2}, \ref{lst:miconic_3} reveal the policy to have correctly identified that boarding a passenger as well as "departing" a passenger at their destination floor are both width-1 subgoals which, together, will reduce the number of unserved passengers. However, these decompositions cannot approximate optimal scheduling with regards to minimal movement between floors, since both subgoals, boarding and departing a passenger, abstract away the $\mathit{up}$/$\mathit{down}$ action. Hence, the plans board all passengers, then deliver them in sequence, as sketch $R3$ would. Interestingly, we never observe sketch $R2$ in this domain either, despite being an equally efficient decomposition in terms of subgoals and primitive actions.   

\paragraph{Reward}
Listings \ref{lst:reward_1} and \ref{lst:reward_2} highlight plans in Reward. Picking a reward is a width-1 subgoal. As such the model simply schedules pickup subgoals in arbitrary sequence without regarding their distance. This comes at no surprise though, since movement, and as such also distance, is fully abstracted away by width-1 subgoals of picking a reward.

\paragraph{Spanner}
The generated plans show that the decompositions follow sketch $R3$ in all cases in which the number of spanners equals the number of nuts to tighten (listing \ref{lst:spanner_1}). Each tightening of a nut requires exactly one non-reuseable spanner. However, in problems with surplus spanners the policy deviates slightly from the associated sketch in two ways. On the one hand, listing \ref{lst:spanner_2} shows the policy choosing the chain of two $\mathit{walk}$ actions as first subgoal, followed by a subgoal of the single action $\mathit{pickup\_spanner}$. Yet, the subgoal of picking up a spanner has width-1 and would subsume both $\mathit{walk}$ actions if selected as the first subgoal, thereby reducing the subgoal plan length by 1. On the other hand, the policy decides to pickup all spanners but one or two in our surplus spanner test-cases (listing \ref{lst:spanner_3}). Explaining these deviations is not readily possible. Firstly, choosing movement-only subgoals is an inefficiency which we conjecture to be a learning artefact. Secondly, since we enforce minimal subgoal plan lengths by virtue of applying reinforcement learning, we would expect the policy to encode the exact difference between surplus spanners and required spanners as feature. The fact that we do not observe the latter gives rise to 2 possible reasons: 1) the model has not seen enough diversity in training instances with various surplus numbers to require encoding this feature, and 2) computing this exact feature is beyond the expressivity of the architecture. 
Since the differences in surplus spanners during training and validation covered between 1 to 3 and our validation ratio converged to 1.0, we cannot assume 1) to be a sufficient explanation. Whether 2) explains the deviations requires further investigation in future work.

\paragraph{Blocksworld}
Blocksworld plans for single tower goals are shown in listing \ref{lst:blocks_single_1}, \ref{lst:blocks_single_2}, and \ref{lst:blocks_single_3}, \ref{lst:blocks_single_4} for width-1 and width-2 subgoals, respectively. Multiple tower goal plans are shown in listing \ref{lst:blocks_multiple_1}, \ref{lst:blocks_multiple_2} (width-1), and \ref{lst:blocks_multiple_3}, \ref{lst:blocks_multiple_4} (width-2). The plans are shown for a problem in each setting once with 10 and once with 20 blocks. We can observe that width-1 policies choose more subgoals than $\mathit{on}$-atoms in the goal, while width-2 is able to choose fewer. The plans also amount to primitive action chains of nearly the same total length for either width, except for one case: the permutation of a 20 block tower (listing \ref{lst:blocks_single_2}). Here, the width-2 plan is twice as long as the width-1 plan, yet achieves this plan with 9 subgoals versus 30. This instance shows the cost that width-2 abstractions may incur: despite solving 19 $\mathit{on}$ atoms in only 9 subgoals, the paths to achieve these subgoals waste actions by building side-towers that need deconstructing during the execution of subsequent subgoals. A less abstracted width-1 path can prevent such artefacts better, as it decomposes $\mathit{on}$ goals into further subproblems. In consequence, width-2 plans generally increase the primitive plan length over width-1 plans.

\clearpage
\begin{figure*}
    \begin{center}
    \huge \textbf{Structure of \siwk Plans}    
    \end{center}
    Subgoals in our traces are represented as chains of actions applied along the \iw{k} path that was found from the source state $s$ and selected by the policy. Each \textit{numbered} line starts a new subgoal. Other linebreaks merely continue the current subgoal's action chain. The actions are concatenated by the separator "\texttt{->}" to indicate the order in which they are enacted. Applying all actions in a subgoal results in a state on which the next subgoal's first action is applied. 
    Only the first plan output of each domain is preceded with a description of the domain representation. Afterwards we provide merely the object names, followed by the initial state and goal description in terms of grounded atoms. Additionally, we highlight for every problem the action-names in orange and the objects in the problem in blue.
\end{figure*}

\lstset{
    basicstyle=\footnotesize\ttfamily,
    keywordstyle=\bfseries, 
    morekeywords=[1]{Domain},
    keywordstyle=[1]\color{red}\bfseries,
    morekeywords=[2]{Objects, Primitive, Goal, Initial, Plan, plan},
    keywordstyle=[2]\bfseries,
    morekeywords=[3]{move, pick\_package, drop\_package},
    keywordstyle=[3]\color{orange}\bfseries,
    morekeywords=[4]{c\_0\_0, c\_0\_1, c\_0\_2, c\_0\_3, c\_0\_4, c\_0\_5, c\_0\_6, c\_0\_7, c\_0\_8, c\_0\_9,
c\_1\_0, c\_1\_1, c\_1\_2, c\_1\_3, c\_1\_4, c\_1\_5, c\_1\_6, c\_1\_7, c\_1\_8, c\_1\_9,
c\_2\_0, c\_2\_1, c\_2\_2, c\_2\_3, c\_2\_4, c\_2\_5, c\_2\_6, c\_2\_7, c\_2\_8, c\_2\_9,
c\_3\_0, c\_3\_1, c\_3\_2, c\_3\_3, c\_3\_4, c\_3\_5, c\_3\_6, c\_3\_7, c\_3\_8, c\_3\_9,
c\_4\_0, c\_4\_1, c\_4\_2, c\_4\_3, c\_4\_4, c\_4\_5, c\_4\_6, c\_4\_7, c\_4\_8, c\_4\_9,
c\_5\_0, c\_5\_1, c\_5\_2, c\_5\_3, c\_5\_4, c\_5\_5, c\_5\_6, c\_5\_7, c\_5\_8, c\_5\_9,
c\_6\_0, c\_6\_1, c\_6\_2, c\_6\_3, c\_6\_4, c\_6\_5, c\_6\_6, c\_6\_7, c\_6\_8, c\_6\_9,
c\_7\_0, c\_7\_1, c\_7\_2, c\_7\_3, c\_7\_4, c\_7\_5, c\_7\_6, c\_7\_7, c\_7\_8, c\_7\_9,
c\_8\_0, c\_8\_1, c\_8\_2, c\_8\_3, c\_8\_4, c\_8\_5, c\_8\_6, c\_8\_7, c\_8\_8, c\_8\_9,
c\_9\_0, c\_9\_1, c\_9\_2, c\_9\_3, c\_9\_4, c\_9\_5, c\_9\_6, c\_9\_7, c\_9\_8, c\_9\_9, p1, p2, p3,p4,p5,p6,p7,p8,p9,p10, t0,t1,t2,t3,t4,t5},
    keywordstyle=[4]\color{blue}\bfseries,
    morekeywords=[5]{at, adjacent, empty},
    keywordstyle=[5]\itshape,
}

\begin{figure*}
\footnotesize
\begin{lstlisting}[basicstyle=\footnotesize\ttfamily]
Domain: delivery
Predicates: adjacent/2, at/2, carrying/2, empty/1
Action schemas:
pick_package(?t: truck, ?p: package, ?x: cell)
pre:  +at(?p, ?x),  +at(?t, ?x),  +empty(?t)
eff:  -at(?p, ?x),  -empty(?t),  +carrying(?t, ?p)
drop_package(?t: truck, ?p: package, ?x: cell)
pre:  +at(?t, ?x),  +carrying(?t, ?p)
eff:  +empty(?t),  -carrying(?t, ?p),  +at(?p, ?x)
move(?t: truck, ?from: cell, ?to: cell)
pre:  +adjacent(?from, ?to),  +at(?t, ?from)
eff:  -at(?t, ?from),  +at(?t, ?to)

Name: delivery_5x5_p-4 (instance_5x5_p-4_0.pddl)
Objects: 
    c_0_0, c_0_1, ..., c_4_3, c_4_4, p1, p2, p3, p4, t1
Initial: 
    at(p4, c_0_0), at(p3, c_0_3), at(p1, c_2_3), at(t1, c_4_1), at(p2, c_4_4)
    empty(t1), adjacent(c_0_1, c_0_0), adjacent(c_1_0, c_0_0), adjacent(c_0_0, c_0_1),
    ..., adjacent(c_4_4, c_4_3), adjacent(c_3_4, c_4_4), adjacent(c_4_3, c_4_4)
Goal: 
    at(p1, c_1_1), at(p2, c_1_1), at(p3, c_1_1), at(p4, c_1_1)

Primitive plan: 39
Plan: 8
1 move(t1, c_4_1, c_3_1) -> move(t1, c_3_1, c_3_0) -> move(t1, c_3_0, c_2_0)
   ->move(t1, c_2_0, c_1_0) -> move(t1, c_1_0, c_0_0) -> pick_package(t1, p4, c_0_0)
2 move(t1, c_0_0, c_1_0) -> move(t1, c_1_0, c_1_1) -> drop_package(t1, p4, c_1_1)
3 move(t1, c_1_1, c_1_2) -> move(t1, c_1_2, c_1_3) -> move(t1, c_1_3, c_1_4)
   ->move(t1, c_1_4, c_2_4) -> move(t1, c_2_4, c_3_4) -> move(t1, c_3_4, c_4_4)
   ->pick_package(t1, p2, c_4_4)
4 move(t1, c_4_4, c_3_4) -> move(t1, c_3_4, c_3_3) -> move(t1, c_3_3, c_3_2)
   ->move(t1, c_3_2, c_3_1) -> move(t1, c_3_1, c_2_1) -> move(t1, c_2_1, c_1_1)
   ->drop_package(t1, p2, c_1_1)
5 move(t1, c_1_1, c_0_1) -> move(t1, c_0_1, c_0_2) -> move(t1, c_0_2, c_0_3)
   ->pick_package(t1, p3, c_0_3)
6 move(t1, c_0_3, c_0_2) -> move(t1, c_0_2, c_1_2) -> move(t1, c_1_2, c_1_1)
   ->drop_package(t1, p3, c_1_1)
7 move(t1, c_1_1, c_1_2) -> move(t1, c_1_2, c_1_3) -> move(t1, c_1_3, c_2_3)
   ->pick_package(t1, p1, c_2_3)
8 move(t1, c_2_3, c_2_2) -> move(t1, c_2_2, c_2_1) -> move(t1, c_2_1, c_1_1)
   ->drop_package(t1, p1, c_1_1)
\end{lstlisting}
\listingcaption{Generated \siwone\ plan for Delivery with $5 \times 5$ grid size, 4 packages, and 1 agent.}
\label{lst:delivery_2}
\end{figure*}

\begin{figure*}
\footnotesize
\begin{lstlisting}[basicstyle=\footnotesize\ttfamily]
Name: delivery-3x9-p_8-t_1 (instance_3x9_p-8.pddl)
Objects: 
    c_0_0, c_0_1, c_0_2, c_0_3, c_0_4, c_0_5, c_0_6, c_0_7, c_0_8, c_1_0, c_1_1
    c_1_2, c_1_3, c_1_4, c_1_5, c_1_6, c_1_7, c_1_8, c_2_0, c_2_1, c_2_2, c_2_3
    c_2_4, c_2_5, c_2_6, c_2_7, c_2_8, p1, p2, p3, p4, p5, p6, p7, p8, t0
Initial: 
    at(p6, c_0_0), at(p7, c_1_1), at(p1, c_1_4), at(p2, c_1_5), at(p4, c_1_6), at(p3
    c_1_8), at(p5, c_2_0), at(t0, c_2_1), at(p8, c_2_5), empty(t0), 
    adjacent(c_0_1, c_0_0), adjacent(c_1_0, c_0_0), adjacent(c_0_0, c_0_1),
    ...
    adjacent(c_2_8, c_2_7), adjacent(c_1_8, c_2_8), adjacent(c_2_7, c_2_8)
Goal: 
    at(p1, c_0_7), at(p2, c_0_7), at(p3, c_0_7), at(p4, c_0_7), at(p5, c_0_7), at(p6
    c_0_7), at(p7, c_0_7), at(p8, c_0_7)

Primitive plan: 94
Plan: 16
1  move(t0, c_2_1, c_2_2) -> move(t0, c_2_2, c_1_2) -> move(t0, c_1_2, c_1_3)
    ->move(t0, c_1_3, c_1_4) -> move(t0, c_1_4, c_1_5) -> pick_package(t0, p2, c_1_5)
2  move(t0, c_1_5, c_1_6) -> move(t0, c_1_6, c_1_7) -> move(t0, c_1_7, c_0_7)
    ->drop_package(t0, p2, c_0_7)
3  move(t0, c_0_7, c_0_8) -> move(t0, c_0_8, c_1_8) -> pick_package(t0, p3, c_1_8)
4  move(t0, c_1_8, c_0_8) -> move(t0, c_0_8, c_0_7) -> drop_package(t0, p3, c_0_7)
5  move(t0, c_0_7, c_1_7) -> move(t0, c_1_7, c_1_6) -> move(t0, c_1_6, c_1_5)
    ->move(t0, c_1_5, c_1_4) -> move(t0, c_1_4, c_2_4) -> move(t0, c_2_4, c_2_3)
    ->move(t0, c_2_3, c_2_2) -> move(t0, c_2_2, c_2_1) -> move(t0, c_2_1, c_2_0)
    ->pick_package(t0, p5, c_2_0)
6  move(t0, c_2_0, c_1_0) -> move(t0, c_1_0, c_0_0) -> move(t0, c_0_0, c_0_1)
    ->move(t0, c_0_1, c_0_2) -> move(t0, c_0_2, c_0_3) -> move(t0, c_0_3, c_0_4)
    ->move(t0, c_0_4, c_0_5) -> move(t0, c_0_5, c_0_6) -> move(t0, c_0_6, c_0_7)
    ->drop_package(t0, p5, c_0_7)
7  move(t0, c_0_7, c_0_6) -> move(t0, c_0_6, c_0_5) -> move(t0, c_0_5, c_0_4)
    ->move(t0, c_0_4, c_0_3) -> move(t0, c_0_3, c_0_2) -> move(t0, c_0_2, c_0_1)
    ->move(t0, c_0_1, c_0_0) -> pick_package(t0, p6, c_0_0)
8  move(t0, c_0_0, c_0_1) -> move(t0, c_0_1, c_0_2) -> move(t0, c_0_2, c_0_3)
    ->move(t0, c_0_3, c_0_4) -> move(t0, c_0_4, c_0_5) -> move(t0, c_0_5, c_0_6)
    ->move(t0, c_0_6, c_0_7) -> drop_package(t0, p6, c_0_7)
9  move(t0, c_0_7, c_1_7) -> move(t0, c_1_7, c_1_6) -> move(t0, c_1_6, c_1_5)
    ->move(t0, c_1_5, c_2_5) -> pick_package(t0, p8, c_2_5)
10 move(t0, c_2_5, c_2_6) -> move(t0, c_2_6, c_1_6) -> move(t0, c_1_6, c_1_7)
    ->move(t0, c_1_7, c_0_7) -> drop_package(t0, p8, c_0_7)
11 move(t0, c_0_7, c_1_7) -> move(t0, c_1_7, c_1_6) -> pick_package(t0, p4, c_1_6)
12 move(t0, c_1_6, c_1_7) -> move(t0, c_1_7, c_0_7) -> drop_package(t0, p4, c_0_7)
13 move(t0, c_0_7, c_1_7) -> move(t0, c_1_7, c_1_6) -> move(t0, c_1_6, c_1_5)
    ->move(t0, c_1_5, c_1_4) -> move(t0, c_1_4, c_1_3) -> move(t0, c_1_3, c_1_2)
    ->move(t0, c_1_2, c_1_1) -> pick_package(t0, p7, c_1_1)
14 move(t0, c_1_1, c_0_1) -> move(t0, c_0_1, c_0_2) -> move(t0, c_0_2, c_0_3)
    ->move(t0, c_0_3, c_0_4) -> move(t0, c_0_4, c_0_5) -> move(t0, c_0_5, c_0_6)
    ->move(t0, c_0_6, c_0_7) -> drop_package(t0, p7, c_0_7)
15 move(t0, c_0_7, c_1_7) -> move(t0, c_1_7, c_1_6) -> move(t0, c_1_6, c_1_5)
    ->move(t0, c_1_5, c_1_4) -> pick_package(t0, p1, c_1_4)
16 move(t0, c_1_4, c_0_4) -> move(t0, c_0_4, c_0_5) -> move(t0, c_0_5, c_0_6)
    ->move(t0, c_0_6, c_0_7) -> drop_package(t0, p1, c_0_7)
\end{lstlisting}
\listingcaption{Generated \siwone\ plan for Delivery with $3 \times 9$ grid size, 8 packages, and 1 agent.}
\label{lst:delivery_3}
\end{figure*}

\lstset{
    basicstyle=\footnotesize\ttfamily,
    keywordstyle=\bfseries, 
    morekeywords=[1]{Domain},
    keywordstyle=[1]\color{red}\bfseries,
    morekeywords=[2]{Objects, Primitive, Goal, Initial, Plan, plan},
    keywordstyle=[2]\bfseries,
    morekeywords=[3]{move, pick, drop },
    keywordstyle=[3]\color{orange}\bfseries,
    morekeywords=[4]{ball19, ball6, ball8, ball14, roomb, ball17, ball20, ball11, ball16, ball21, ball3, ball1, ball18, ball5, ball15, ball13, ball12, ball2, ball4, ball10, right, ball7, ball9,
ball22, ball23, ball24, ball25, ball26, ball27, ball28, ball29, ball30, ball31, ball32, ball33, ball34, ball35, ball36, ball37, ball38, ball39, ball40, ball41, ball42, ball43, ball44, ball45, ball46, ball47, ball48, ball49, ball50, ball51, ball52, ball53, ball54, ball55, ball56, ball57, ball58, ball59, ball60, ball61, rooma, left, ball22 },
    keywordstyle=[4]\color{blue}\bfseries,
}

\begin{figure*}
\footnotesize
\begin{lstlisting}[basicstyle=\footnotesize\ttfamily]
Domain: gripper
Types: object
Predicates: at/2, at-robby/1, ball/1, carry/2, free/1, gripper/1, room/1
Action schemas:
move(?from: object, ?to: object)
pre:  +room(?from),  +room(?to),  +at-robby(?from)
eff:  +at-robby(?to),  -at-robby(?from)
pick(?obj: object, ?room: object, ?gripper: object)
pre:  +ball(?obj),  +room(?room),  +gripper(?gripper),  +at(?obj, ?room),  +at-robby(?room),  +free(?gripper)
eff:  +carry(?obj, ?gripper),  -at(?obj, ?room),  -free(?gripper)
drop(?obj: object, ?room: object, ?gripper: object)
pre:  +ball(?obj),  +room(?room),  +gripper(?gripper),  +carry(?obj, ?gripper),  +at-robby(?room)
eff:  +at(?obj, ?room),  +free(?gripper),  -carry(?obj, ?gripper)

Name: gripper-22 (gripper_b-22.pddl)
Objects: 
    ball1, ..., ball22, left, right, rooma, roomb
Initial: 
    room(rooma), room(roomb), ball(ball1), ball(ball2), ..., ball(ball21), ball(ball22),
    gripper(left), gripper(right), at-robby(rooma), at(ball1, rooma), at(ball2, rooma), 
    ..., at(ball21, rooma), at(ball22, rooma), free(left), free(right)
Goal: 
    at(ball1, roomb), at(ball2, roomb), ..., at(ball21, roomb), at(ball22, roomb)

Primitive plan: 85
Plan: 44
1  pick(ball16, rooma, right)
2  pick(ball21, rooma, left)
3  move(rooma, roomb) -> drop(ball16, roomb, right)
4  move(roomb, rooma) -> pick(ball3, rooma, right)
5  move(rooma, roomb) -> drop(ball3, roomb, right)
6  move(roomb, rooma) -> pick(ball10, rooma, right)
7  move(rooma, roomb) -> drop(ball10, roomb, right)
8  move(roomb, rooma) -> pick(ball1, rooma, right)
9  move(rooma, roomb) -> drop(ball21, roomb, left)
...
39 move(rooma, roomb) -> drop(ball11, roomb, right)
40 move(roomb, rooma) -> pick(ball6, rooma, right)
41 move(rooma, roomb) -> drop(ball2, roomb, left)
42 move(roomb, rooma) -> pick(ball18, rooma, left)
43 move(rooma, roomb) -> drop(ball6, roomb, right)
44 drop(ball18, roomb, left
\end{lstlisting}
\listingcaption{Generated \siwone\ plan for Gripper with $22$ balls.}
\label{lst:gripper_1}
\end{figure*}

\begin{figure*}
\footnotesize
\begin{lstlisting}[basicstyle=\footnotesize\ttfamily]
Name: gripper-55 (gripper_b-55.pddl)
Objects: 
    ball1, ..., ball55, left, right, rooma, roomb
Initial: 
    room(rooma), room(roomb), ball(ball1), ball(ball2), ..., ball(ball55)
    gripper(left), gripper(right), at-robby(rooma), at(ball1, rooma), at(ball2
    rooma), at(ball3, rooma), ..., at(ball54, rooma), at(ball55, rooma),
    free(left), free(right)
Goal: 
    at(ball1, roomb), at(ball2, roomb), ..., at(ball54, roomb), at(ball55, roomb)

Primitive plan: 217
Plan: 110
1   pick(ball33, rooma, right)
2   pick(ball46, rooma, left)
3   move(rooma, roomb) -> drop(ball46, roomb, left)
4   move(roomb, rooma) -> pick(ball37, rooma, left)
5   move(rooma, roomb) -> drop(ball37, roomb, left)
6   move(roomb, rooma) -> pick(ball15, rooma, left)
7   move(rooma, roomb) -> drop(ball33, roomb, right)
8   move(roomb, rooma) -> pick(ball50, rooma, right)
9   move(rooma, roomb) -> drop(ball50, roomb, right)
10  move(roomb, rooma) -> pick(ball1, rooma, right)
11  move(rooma, roomb) -> drop(ball1, roomb, right)
12  move(roomb, rooma) -> pick(ball24, rooma, right)
13  move(rooma, roomb) -> drop(ball15, roomb, left)
14  move(roomb, rooma) -> pick(ball43, rooma, left)
15  move(rooma, roomb) -> drop(ball43, roomb, left)
...
94  move(roomb, rooma) -> pick(ball42, rooma, left)
95  move(rooma, roomb) -> drop(ball42, roomb, left)
96  move(roomb, rooma) -> pick(ball19, rooma, left)
97  move(rooma, roomb) -> drop(ball19, roomb, left)
98  move(roomb, rooma) -> pick(ball9, rooma, left)
99  move(rooma, roomb) -> drop(ball48, roomb, right)
100 move(roomb, rooma) -> pick(ball14, rooma, right)
101 move(rooma, roomb) -> drop(ball9, roomb, left)
102 move(roomb, rooma) -> pick(ball7, rooma, left)
103 move(rooma, roomb) -> drop(ball7, roomb, left)
104 move(roomb, rooma) -> pick(ball41, rooma, left)
105 move(rooma, roomb) -> drop(ball14, roomb, right)
106 move(roomb, rooma) -> pick(ball11, rooma, right)
107 move(rooma, roomb) -> drop(ball41, roomb, left)
108 move(roomb, rooma) -> pick(ball54, rooma, left)
109 move(rooma, roomb) -> drop(ball11, roomb, right)
110 drop(ball54, roomb, left)
\end{lstlisting}
\listingcaption{Generated \siwone\ plan for Gripper with $55$ balls.}
\label{lst:gripper_3}
\end{figure*}

\lstset{
    basicstyle=\footnotesize\ttfamily,
    keywordstyle=\bfseries, 
    language=,
    morekeywords=[1]{Domain},
    keywordstyle=[1]\color{red}\bfseries,
    morekeywords=[2]{Objects, Primitive, Goal, Initial, Plan, plan},
    keywordstyle=[2]\bfseries,
    morekeywords=[3]{make_sandwich_no_gluten, make_sandwich, serve_sandwich, serve_sandwich_no_gluten, move_tray, put_on_tray},
    keywordstyle=[3]\color{orange}\bfseries,
    morekeywords=[4]{} 
}

\begin{figure*}
\footnotesize
\begin{lstlisting}[basicstyle=\footnotesize\ttfamily]
Domain: child-snack
Types: bread-portion, child, content-portion, object, place, sandwich, tray
       Predicates: allergic_gluten/1, at/2, at_kitchen_bread/1, at_kitchen_content/1, 
       at_kitchen_sandwich/1, no_gluten_bread/1, no_gluten_content/1, no_gluten_sandwich/1, 
       not_allergic_gluten/1, notexist/1, ontray/2, served/1, waiting/2
Action schemas:
make_sandwich_no_gluten(?s: sandwich, ?b: bread-portion, ?c: content-portion)
pre:  +at_kitchen_bread(?b),  +at_kitchen_content(?c),  +no_gluten_bread(?b),  +no_gluten_content(?c),  +notexist(?s)
eff:  -at_kitchen_bread(?b),  -at_kitchen_content(?c),  +at_kitchen_sandwich(?s),  +no_gluten_sandwich(?s),  -notexist(?s)
make_sandwich(?s: sandwich, ?b: bread-portion, ?c: content-portion)
pre:  +at_kitchen_bread(?b),  +at_kitchen_content(?c),  +notexist(?s)
eff:  -at_kitchen_bread(?b),  -at_kitchen_content(?c),  +at_kitchen_sandwich(?s),  -notexist(?s)
put_on_tray(?s: sandwich, ?t: tray)
pre:  +at_kitchen_sandwich(?s),  +at(?t, kitchen)
eff:  -at_kitchen_sandwich(?s),  +ontray(?s, ?t)
serve_sandwich_no_gluten(?s: sandwich, ?c: child, ?t: tray, ?p: place)
pre:  +allergic_gluten(?c),  +ontray(?s, ?t),  +waiting(?c, ?p),  +no_gluten_sandwich(?s),  +at(?t, ?p)
eff:  -ontray(?s, ?t),  +served(?c)
serve_sandwich(?s: sandwich, ?c: child, ?t: tray, ?p: place)
pre:  +not_allergic_gluten(?c),  +waiting(?c, ?p),  +ontray(?s, ?t),  +at(?t, ?p)
eff:  -ontray(?s, ?t),  +served(?c)
move_tray(?t: tray, ?p1: place, ?p2: place)
pre:  +at(?t, ?p1)
eff:  -at(?t, ?p1),  +at(?t, ?p2)
\end{lstlisting}

\lstset{
    basicstyle=\footnotesize\ttfamily,
    keywordstyle=\bfseries, 
    morekeywords=[1]{Domain},
    keywordstyle=[1]\color{red}\bfseries,
    morekeywords=[2]{Objects, Primitive, Goal, Initial, Plan, plan},
    keywordstyle=[2]\bfseries,
    morekeywords=[3]{make_sandwich_no_gluten, make_sandwich, serve_sandwich, serve_sandwich_no_gluten, move_tray, put_on_tray },
    keywordstyle=[3]\color{orange}\bfseries,
    morekeywords=[4]{table1, bread1,bread2,bread3,bread4,table3, tray1,tray2,tray3, content1,content2,content3, content4, sandw1,sandw2,sandw3,sandw4, child1,child2,child3,child4, table2, kitchen },
    keywordstyle=[4]\color{blue}\bfseries,
}

\begin{lstlisting}[basicstyle=\footnotesize\ttfamily]
Name: prob-snack (p-1-1.0-0.0-1-5.pddl)
Objects: 
    bread1, child1, content1, sandw1, table1, table2, table3, tray1, kitchen
Initial: 
    at_kitchen_bread(bread1), at_kitchen_content(content1)
    not_allergic_gluten(child1), waiting(child1, table3), at(tray1, kitchen)
    notexist(sandw1)
Goal: 
    served(child1)

Primitive plan: 4
Plan: 2
1 make_sandwich(sandw1, bread1, content1) -> put_on_tray(sandw1, tray1)
2 move_tray(tray1, kitchen, table3) -> serve_sandwich(sandw1, child1, tray1, table3)
\end{lstlisting}
\listingcaption{Generated \siwone\ plan for Childsnack with 1 not gluten-allergic child.}
\label{lst:childsnack_1}
\end{figure*}

       \lstset{
    basicstyle=\footnotesize\ttfamily,
    keywordstyle=\bfseries, 
    morekeywords=[1]{Domain},
    keywordstyle=[1]\color{red}\bfseries,
    morekeywords=[2]{Objects, Primitive, Goal, Initial, Plan, plan},
    keywordstyle=[2]\bfseries,
    morekeywords=[3]{make_sandwich_no_gluten, make_sandwich, serve_sandwich, serve_sandwich_no_gluten, move_tray, put_on_tray },
    keywordstyle=[3]\color{orange}\bfseries,
    morekeywords=[4]{table1, bread1,bread2,bread3,bread4,table3, tray1,tray2,tray3, content1,content2,content3, content4, sandw1,sandw2,sandw3,sandw4, child1,child2,child3,child4, table2, kitchen },
    keywordstyle=[4]\color{blue}\bfseries,
}     
\begin{figure*}
\footnotesize
\begin{lstlisting}[basicstyle=\footnotesize\ttfamily]
Name: prob-snack (p-1-1.0-1.0-1-6.pddl)
Objects: 
    bread1, child1, content1, sandw1, table1, table2, table3, tray1, kitchen
Initial: 
    at_kitchen_bread(bread1), at_kitchen_content(content1), no_gluten_bread(bread1)
    no_gluten_content(content1), allergic_gluten(child1), waiting(child1, table1)
    at(tray1, kitchen), notexist(sandw1)
Goal: 
    served(child1)

Primitive plan: 4
Plan: 2
1 make_sandwich_no_gluten(sandw1, bread1, content1) -> put_on_tray(sandw1, tray1)
2 move_tray(tray1, kitchen, table1) -> serve_sandwich_no_gluten(sandw1, child1, tray1, table1)
\end{lstlisting}
\listingcaption{Generated \siwone\ plan for Childsnack with 1 gluten-allergic child.}
\label{lst:childsnack_2}
\end{figure*}

\begin{figure*}
\footnotesize
\begin{lstlisting}[basicstyle=\footnotesize\ttfamily]
Name: prob-snack (p-2-1.0-0.5-1-4.pddl)
Objects: 
    bread1, bread2, child1, child2, content1, content2, sandw1, sandw2, table1
    table2, table3, tray1, kitchen
Initial: 
    at_kitchen_bread(bread1), at_kitchen_bread(bread2), at_kitchen_content(content1)
    at_kitchen_content(content2), no_gluten_bread(bread1)
    no_gluten_content(content2), allergic_gluten(child1)
    not_allergic_gluten(child2), waiting(child2, table2), waiting(child1, table3)
    at(tray1, kitchen), notexist(sandw1), notexist(sandw2)
Goal: 
    served(child1), served(child2)

Primitive plan: 8
Plan: 4
1 make_sandwich_no_gluten(sandw2, bread1, content2) -> put_on_tray(sandw2, tray1)
2 make_sandwich(sandw1, bread2, content1) -> put_on_tray(sandw1, tray1)
3 move_tray(tray1, kitchen, table3) -> serve_sandwich_no_gluten(sandw2, child1, tray1, table3)
4 move_tray(tray1, table3, table2) -> serve_sandwich(sandw1, child2, tray1, table2)
\end{lstlisting}
\listingcaption{Generated \siwone\ plan for Childsnack with 1 gluten-allergic child and 1 non-allergic child.}
\end{figure*}
            
\begin{figure*}
\footnotesize
\begin{lstlisting}[basicstyle=\footnotesize\ttfamily]
Name: prob-snack (p-3-1.0-0.33-1-1.pddl)
Objects: 
    bread1, bread2, bread3, child1, child2, child3, content1, content2, content3
    sandw1, sandw2, sandw3, table1, table2, table3, tray1, kitchen
Initial: 
    at_kitchen_bread(bread1), at_kitchen_bread(bread2), at_kitchen_bread(bread3)
    at_kitchen_content(content1), at_kitchen_content(content2)
    at_kitchen_content(content3), no_gluten_bread(bread1)
    no_gluten_content(content3), allergic_gluten(child1)
    not_allergic_gluten(child2), not_allergic_gluten(child3), waiting(child2
    table1), waiting(child1, table2), waiting(child3, table2), at(tray1, kitchen)
    notexist(sandw1), notexist(sandw2), notexist(sandw3)
Goal: 
    served(child1), served(child2), served(child3)

Primitive plan: 12
Plan: 6
1 make_sandwich_no_gluten(sandw1, bread1, content3) -> put_on_tray(sandw1, tray1)
2 make_sandwich(sandw2, bread2, content1) -> put_on_tray(sandw2, tray1)
3 make_sandwich(sandw3, bread3, content2) -> put_on_tray(sandw3, tray1)
4 move_tray(tray1, kitchen, table2) -> serve_sandwich_no_gluten(sandw1, child1, tray1, table2)
5 move_tray(tray1, table2, table1) -> serve_sandwich(sandw2, child2, tray1, table1)
6 move_tray(tray1, table1, table2) -> serve_sandwich(sandw3, child3, tray1, table2)
\end{lstlisting}
\listingcaption{Generated \siwone\ plan for Childsnack with 1 gluten-allergic child and 2 non-allergic children.}
\label{lst:childsnack_3}
\end{figure*}
            
\begin{figure*}
\footnotesize
\begin{lstlisting}[basicstyle=\footnotesize\ttfamily]
Name: prob-snack (p-4-1.0-0.48-1-1.pddl)
Objects: 
    bread1, ..., bread4, child1, ..., child4, content1
    ..., content4, sandw1, ..., sandw4, table1, table2,
    table3, tray1, kitchen
Initial: 
    at_kitchen_bread(bread1), ...,
    at_kitchen_bread(bread4), at_kitchen_content(content1), ...
    at_kitchen_content(content4), no_gluten_bread(bread4)
    no_gluten_content(content4), allergic_gluten(child2)
    not_allergic_gluten(child1), not_allergic_gluten(child3)
    not_allergic_gluten(child4), waiting(child1, table2), waiting(child2, table3)
    waiting(child3, table3), waiting(child4, table3), at(tray1, kitchen)
    notexist(sandw1), notexist(sandw2), notexist(sandw3), notexist(sandw4)
Goal: 
    served(child1), served(child2), served(child3), served(child4)

Primitive plan: 14
Plan: 8
1 make_sandwich_no_gluten(sandw2, bread4, content4) -> put_on_tray(sandw2, tray1)
2 make_sandwich(sandw4, bread3, content2) -> put_on_tray(sandw4, tray1)
3 make_sandwich(sandw1, bread2, content1) -> put_on_tray(sandw1, tray1)
4 make_sandwich(sandw3, bread1, content3) -> put_on_tray(sandw3, tray1)
5 move_tray(tray1, kitchen, table2) -> serve_sandwich(sandw3, child1, tray1, table2)
6 move_tray(tray1, table2, table3) -> serve_sandwich_no_gluten(sandw2, child2, tray1, table3)
7 serve_sandwich(sandw4, child3, tray1, table3)
8 serve_sandwich(sandw1, child4, tray1, table3
\end{lstlisting}
\listingcaption{Generated \siwone\ plan for Childsnack with 1 gluten-allergic child and 3 non-allergic children.}
\label{lst:childsnack_4}
\end{figure*}

\lstset{
    basicstyle=\footnotesize\ttfamily,
    keywordstyle=\bfseries, 
    morekeywords=[1]{Domain},
    keywordstyle=[1]\color{red}\bfseries,
    morekeywords=[2]{Objects, Primitive, Goal, Initial, Plan, plan},
    keywordstyle=[2]\bfseries,
    morekeywords=[3]{depart, down, up, board },
    keywordstyle=[3]\color{orange}\bfseries
}
            
\begin{figure*}
\footnotesize
\begin{lstlisting}[basicstyle=\footnotesize\ttfamily]
Domain: miconic
Types: object
Predicates: above/2, boarded/1, destin/2, floor/1, lift-at/1, origin/2, passenger/1, served/1
Action schemas:
board(?f: object, ?p: object)
pre:  +floor(?f),  +passenger(?p),  +lift-at(?f),  +origin(?p, ?f)
eff:  +boarded(?p)
depart(?f: object, ?p: object)
pre:  +floor(?f),  +passenger(?p),  +lift-at(?f),  +destin(?p, ?f),  +boarded(?p)
eff:  -boarded(?p),  +served(?p)
up(?f1: object, ?f2: object)
pre:  +floor(?f1),  +floor(?f2),  +lift-at(?f1),  +above(?f1, ?f2)
eff:  +lift-at(?f2),  -lift-at(?f1)
down(?f1: object, ?f2: object)
pre:  +floor(?f1),  +floor(?f2),  +lift-at(?f1),  +above(?f2, ?f1)
eff:  +lift-at(?f2),  -lift-at(?f1)

\end{lstlisting}
\lstset{
    basicstyle=\footnotesize\ttfamily,
    keywordstyle=\bfseries, 
    morekeywords=[1]{Domain},
    keywordstyle=[1]\color{red}\bfseries,
    morekeywords=[2]{Objects, Primitive, Goal, Initial, Plan, plan},
    keywordstyle=[2]\bfseries,
    morekeywords=[3]{depart, down, up, board },
    keywordstyle=[3]\color{orange}\bfseries,
    morekeywords=[4]{p12, p1, f13, f35, f46, p10, f48, p20, f19, f40, f59, f32, p27, p2, f26, f29, f22, f33, f56, f5, p0, f38, p7, f18, p25, f42, p16, f41, p13, f25, f51, p19, f53, f49, f47, f17, f21, f30, f36, f44, p22, p6, f6, f24, f11, p3, f10, f4, f27, p28, f39, f23, f57, f2, p17, f1, p4, p15, p11, f34, p8, p26, f8, f7, f43, p21, f55, p9, f15, f58, f31, p24, p18, f20, f37, f12, f50, p23, f9, f0, f52, f54, f14, f45, p5, f16, f28, p29, f3, p14 },
    keywordstyle=[4]\color{blue}\bfseries,
}  
\begin{lstlisting}

Name: mixed-f2-p1-u0-v0-g0-a0-n0-a0-b0-n0-f0-r0 (s1-0.pddl)
Objects: 
    f0, f1, p0
Initial: 
    origin(p0, f1), floor(f0), floor(f1), passenger(p0), destin(p0, f0), above(f0, f1), lift-at(f0)
Goal: 
    served(p0)

Primitive plan: 4
Plan: 2
1 up(f0, f1) -> board(f1, p0)
2 down(f1, f0) -> depart(f0, p0)
\end{lstlisting}
\listingcaption{Generated \siwone\ plan for Miconic with $1$ passenger and 2 floors.}
\label{lst:miconic_1}
\end{figure*}

\lstset{
    basicstyle=\footnotesize\ttfamily,
    keywordstyle=\bfseries, 
    morekeywords=[1]{Domain},
    keywordstyle=[1]\color{red}\bfseries,
    morekeywords=[2]{Objects, Primitive, Goal, Initial, Plan, plan},
    keywordstyle=[2]\bfseries,
    morekeywords=[3]{depart, down, up, board },
    keywordstyle=[3]\color{orange}\bfseries,
    morekeywords=[4]{p12, p1, f13, f35, f46, p10, f48, p20, f19, f40, f59, f32, p27, p2, f26, f29, f22, f33, f56, f5, p0, f38, p7, f18, p25, f42, p16, f41, p13, f25, f51, p19, f53, f49, f47, f17, f21, f30, f36, f44, p22, p6, f6, f24, f11, p3, f10, f4, f27, p28, f39, f23, f57, f2, p17, f1, p4, p15, p11, f34, p8, p26, f8, f7, f43, p21, f55, p9, f15, f58, f31, p24, p18, f20, f37, f12, f50, p23, f9, f0, f52, f54, f14, f45, p5, f16, f28, p29, f3, p14 },
    keywordstyle=[4]\color{blue}\bfseries,
}
            
\begin{figure*}
\footnotesize
\begin{lstlisting}[basicstyle=\footnotesize\ttfamily]
Name: mixed-f20-p10-u0-v0-g0-a0-n0-a0-b0-n0-f0-r0 (s10-0.pddl)
Objects: 
    f0, ..., f19, p0, ..., p9
Initial: 
    origin(p8, f0), origin(p5, f2), origin(p0, f3), origin(p7, f3), origin(p3, f6)
    origin(p4, f9), origin(p6, f10), origin(p9, f12), origin(p2, f13), origin(p1
    f17), floor(f0), 
    floor(f1), ..., floor(f19), 
    passenger(p0), passenger(p1), ..., passenger(p9), 
    destin(p4, f1), destin(p0, f6), destin(p7, f6), destin(p8, f6)
    destin(p5, f7), destin(p3, f12), destin(p1, f15), destin(p2, f15), destin(p9
    f16), destin(p6, f19),
    above(f0, f1), above(f0, f2), above(f1, f2), , ..., above(f18, f19), lift-at(f0)
Goal: 
    served(p0), served(p1), served(p2), served(p3), served(p4), served(p5)
    served(p6), served(p7), served(p8), served(p9)

Primitive plan: 36
Plan: 20
1  board(f0, p8)
2  up(f0, f2) -> board(f2, p5)
3  up(f2, f9) -> board(f9, p4)
4  up(f9, f10) -> board(f10, p6)
5  up(f10, f13) -> board(f13, p2)
6  up(f13, f17) -> board(f17, p1)
7  down(f17, f3) -> board(f3, p0)
8  board(f3, p7)
9  up(f3, f6) -> board(f6, p3)
10 up(f6, f12) -> board(f12, p9)
11 down(f12, f1) -> depart(f1, p4)
12 up(f1, f7) -> depart(f7, p5)
13 up(f7, f19) -> depart(f19, p6)
14 down(f19, f6) -> depart(f6, p8)
15 up(f6, f16) -> depart(f16, p9)
16 down(f16, f12) -> depart(f12, p3)
17 up(f12, f15) -> depart(f15, p1)
18 depart(f15, p2)
19 down(f15, f6) -> depart(f6, p0)
20 depart(f6, p7)
\end{lstlisting}
\listingcaption{Generated \siwone\ plan for Miconic with $10$ passengers and $20$ floors.}
\label{lst:miconic_2}
\end{figure*}

\lstset{
    basicstyle=\footnotesize\ttfamily,
    keywordstyle=\bfseries, 
    morekeywords=[1]{Domain},
    keywordstyle=[1]\color{red}\bfseries,
    morekeywords=[2]{Objects, Primitive, Goal, Initial, Plan, plan},
    keywordstyle=[2]\bfseries,
    morekeywords=[3]{depart, down, up, board },
    keywordstyle=[3]\color{orange}\bfseries,
    morekeywords=[4]{p3, p12, p1, p13, f10, f13, f25, f4, f27, p9, p10, f15, f19, f32, f23, f17, p2, f21, f26, f30, f31, f2, f20, f29, f1, f12, f22, p4, f9, f0, p15, f33, p11, p6, f5, f6, p0, f24, f14, p7, p5, f16, f28, p8, f11, f8, f18, f3, f7, p14, p16 },
    keywordstyle=[4]\color{blue}\bfseries,
}

\begin{figure*}
\footnotesize
\begin{lstlisting}[basicstyle=\footnotesize\ttfamily]
Name: mixed-f34-p17-u0-v0-g0-a0-n0-a0-b0-n0-f0-r0 (s17-0.pddl)
Objects: 
    f0, ..., f33, p0, ..., p17
Initial: 
    origin(p10, f1), origin(p9, f6), origin(p13, f8), origin(p1, f9), origin(p16
    f10), origin(p4, f11), origin(p12, f14), origin(p2, f15), origin(p15, f17)
    origin(p11, f19), origin(p3, f20), origin(p6, f20), origin(p5, f22), origin(p0
    f27), origin(p8, f28), origin(p7, f29), origin(p14, f31), floor(f0), ..., floor(f33),
    passenger(p0), ..., passenger(p16), destin(p16, f0), destin(p2, f3)
    destin(p11, f5), destin(p10, f8), destin(p14, f11), destin(p3, f14), destin(p9
    f14), destin(p1, f17), destin(p4, f19), destin(p6, f19), destin(p13, f21)
    destin(p7, f22), destin(p12, f22), destin(p8, f30), destin(p5, f31), destin(p0
    f32), destin(p15, f32), 
    above(f0, f1), above(f0, f2), ..., above(f32, f33), lift-at(f0)
Goal: 
    served(p0), served(p1), served(p2), served(p3), served(p4), served(p5)
    served(p6), served(p7), served(p8), served(p9), served(p10), served(p11)
    served(p12), served(p13), served(p14), served(p15), served(p16)

Primitive plan: 63
Plan: 34
1  up(f0, f9) -> board(f9, p1)
2  up(f9, f28) -> board(f28, p8)
3  down(f28, f15) -> board(f15, p2)
4  down(f15, f1) -> board(f1, p10)
5  up(f1, f6) -> board(f6, p9)
6  up(f6, f29) -> board(f29, p7)
7  down(f29, f10) -> board(f10, p16)
8  down(f10, f8) -> board(f8, p13)
9  up(f8, f20) -> board(f20, p3)
10 board(f20, p6)
11 down(f20, f14) -> board(f14, p12)
12 up(f14, f22) -> board(f22, p5)
13 up(f22, f27) -> board(f27, p0)
14 down(f27, f17) -> board(f17, p15)
15 down(f17, f11) -> board(f11, p4)
16 up(f11, f19) -> board(f19, p11)
17 up(f19, f31) -> board(f31, p14)
18 down(f31, f0) -> depart(f0, p16)
19 up(f0, f3) -> depart(f3, p2)
20 up(f3, f5) -> depart(f5, p11)
21 up(f5, f8) -> depart(f8, p10)
22 up(f8, f21) -> depart(f21, p13)
23 down(f21, f17) -> depart(f17, p1)
24 up(f17, f30) -> depart(f30, p8)
25 down(f30, f11) -> depart(f11, p14)
26 up(f11, f31) -> depart(f31, p5)
27 down(f31, f19) -> depart(f19, p4)
28 depart(f19, p6)
29 up(f19, f22) -> depart(f22, p7)
30 depart(f22, p12)
31 down(f22, f14) -> depart(f14, p9)
32 depart(f14, p3)
33 up(f14, f32) -> depart(f32, p15)
34 depart(f32, p0
\end{lstlisting}
\listingcaption{Generated \siwone\ plan for Miconic with $17$ passengers and $34$ floors.}
\label{lst:miconic_3}
\end{figure*}

\lstset{
    basicstyle=\footnotesize\ttfamily,
    keywordstyle=\bfseries, 
    morekeywords=[1]{Domain},
    keywordstyle=[1]\color{red}\bfseries,
    morekeywords=[2]{Objects, Primitive, Goal, Initial, Plan, plan},
    keywordstyle=[2]\bfseries,
    morekeywords=[3]{pick, move },
    keywordstyle=[3]\color{orange}\bfseries,
    morekeywords=[4]{c\_0\_0, c\_0\_1, c\_0\_2, c\_0\_3, c\_0\_4, c\_0\_5, c\_0\_6, c\_0\_7, c\_0\_8, c\_0\_9, c\_0\_10, c\_0\_11, c\_0\_12, c\_0\_13, c\_0\_14, c\_0\_15, c\_0\_16,
c\_1\_0, c\_1\_1, c\_1\_2, c\_1\_3, c\_1\_4, c\_1\_5, c\_1\_6, c\_1\_7, c\_1\_8, c\_1\_9, c\_1\_10, c\_1\_11, c\_1\_12, c\_1\_13, c\_1\_14, c\_1\_15, c\_1\_16,
c\_2\_0, c\_2\_1, c\_2\_2, c\_2\_3, c\_2\_4, c\_2\_5, c\_2\_6, c\_2\_7, c\_2\_8, c\_2\_9, c\_2\_10, c\_2\_11, c\_2\_12, c\_2\_13, c\_2\_14, c\_2\_15, c\_2\_16,
c\_3\_0, c\_3\_1, c\_3\_2, c\_3\_3, c\_3\_4, c\_3\_5, c\_3\_6, c\_3\_7, c\_3\_8, c\_3\_9, c\_3\_10, c\_3\_11, c\_3\_12, c\_3\_13, c\_3\_14, c\_3\_15, c\_3\_16,
c\_4\_0, c\_4\_1, c\_4\_2, c\_4\_3, c\_4\_4, c\_4\_5, c\_4\_6, c\_4\_7, c\_4\_8, c\_4\_9, c\_4\_10, c\_4\_11, c\_4\_12, c\_4\_13, c\_4\_14, c\_4\_15, c\_4\_16,
c\_5\_0, c\_5\_1, c\_5\_2, c\_5\_3, c\_5\_4, c\_5\_5, c\_5\_6, c\_5\_7, c\_5\_8, c\_5\_9, c\_5\_10, c\_5\_11, c\_5\_12, c\_5\_13, c\_5\_14, c\_5\_15, c\_5\_16,
c\_6\_0, c\_6\_1, c\_6\_2, c\_6\_3, c\_6\_4, c\_6\_5, c\_6\_6, c\_6\_7, c\_6\_8, c\_6\_9, c\_6\_10, c\_6\_11, c\_6\_12, c\_6\_13, c\_6\_14, c\_6\_15, c\_6\_16,
c\_7\_0, c\_7\_1, c\_7\_2, c\_7\_3, c\_7\_4, c\_7\_5, c\_7\_6, c\_7\_7, c\_7\_8, c\_7\_9, c\_7\_10, c\_7\_11, c\_7\_12, c\_7\_13, c\_7\_14, c\_7\_15, c\_7\_16,
c\_8\_0, c\_8\_1, c\_8\_2, c\_8\_3, c\_8\_4, c\_8\_5, c\_8\_6, c\_8\_7, c\_8\_8, c\_8\_9, c\_8\_10, c\_8\_11, c\_8\_12, c\_8\_13, c\_8\_14, c\_8\_15, c\_8\_16,
c\_9\_0, c\_9\_1, c\_9\_2, c\_9\_3, c\_9\_4, c\_9\_5, c\_9\_6, c\_9\_7, c\_9\_8, c\_9\_9, c\_9\_10, c\_9\_11, c\_9\_12, c\_9\_13, c\_9\_14, c\_9\_15, c\_9\_16,
c\_10\_0, c\_10\_1, c\_10\_2, c\_10\_3, c\_10\_4, c\_10\_5, c\_10\_6, c\_10\_7, c\_10\_8, c\_10\_9, c\_10\_10, c\_10\_11, c\_10\_12, c\_10\_13, c\_10\_14, c\_10\_15, c\_10\_16,
c\_11\_0, c\_11\_1, c\_11\_2, c\_11\_3, c\_11\_4, c\_11\_5, c\_11\_6, c\_11\_7, c\_11\_8, c\_11\_9, c\_11\_10, c\_11\_11, c\_11\_12, c\_11\_13, c\_11\_14, c\_11\_15, c\_11\_16,
c\_12\_0, c\_12\_1, c\_12\_2, c\_12\_3, c\_12\_4, c\_12\_5, c\_12\_6, c\_12\_7, c\_12\_8, c\_12\_9, c\_12\_10, c\_12\_11, c\_12\_12, c\_12\_13, c\_12\_14, c\_12\_15, c\_12\_16,
c\_13\_0, c\_13\_1, c\_13\_2, c\_13\_3, c\_13\_4, c\_13\_5, c\_13\_6, c\_13\_7, c\_13\_8, c\_13\_9, c\_13\_10, c\_13\_11, c\_13\_12, c\_13\_13, c\_13\_14, c\_13\_15, c\_13\_16,
c\_14\_0, c\_14\_1, c\_14\_2, c\_14\_3, c\_14\_4, c\_14\_5, c\_14\_6, c\_14\_7, c\_14\_8, c\_14\_9, c\_14\_10, c\_14\_11, c\_14\_12, c\_14\_13, c\_14_14, c\_14\_15, c\_14\_16,
c\_15\_0, c\_15\_1, c\_15\_2, c\_15\_3, c\_15\_4, c\_15\_5, c\_15\_6, c\_15\_7, c\_15\_8, c\_15\_9, c\_15\_10, c\_15\_11, c\_15\_12, c\_15\_13, c\_15\_14, c\_15\_15, c\_15\_16,
c\_16\_0, c\_16\_1, c\_16\_2, c\_16\_3, c\_16\_4, c\_16\_5, c\_16\_6, c\_16\_7, c\_16\_8, c\_16\_9, c\_16\_10, c\_16\_11, c\_16\_12, c\_16\_13, c\_16\_14, c\_16\_15, c\_16\_16},
    keywordstyle=[4]\color{blue}\bfseries,
}

\begin{figure*}
\footnotesize
\begin{lstlisting}[basicstyle=\footnotesize\ttfamily]
Domain: reward
Types: cell, object
Predicates: adjacent/2, at/1, picked/1, reward/1, unblocked/1
Action schemas:
move(?from: cell, ?to: cell)
pre:  +adjacent(?from, ?to),  +at(?from),  +unblocked(?to)
eff:  -at(?from),  +at(?to)
pick-reward(?x: cell)
pre:  +at(?x),  +reward(?x)
eff:  -reward(?x),  +picked(?x)

Name: reward-10x10 (instance_10x10_0.pddl)
Objects: 
    c_0_0, ..., c_9_9
Initial: 
    at(c_0_0), reward(c_0_1), reward(c_0_5), reward(c_3_8), reward(c_4_5)
    reward(c_4_9), reward(c_6_1), reward(c_7_5), reward(c_8_5), unblocked(c_0_0)
    unblocked(c_0_1), ..., unblocked(c_9_9), adjacent(c_0_1, c_0_0)
    adjacent(c_1_0, c_0_0), adjacent(c_0_0, c_0_1), ...., adjacent(c_9_9, c_9_8)
    adjacent(c_8_9, c_9_9), adjacent(c_9_8, c_9_9)
Goal: 
    picked(c_0_5), picked(c_8_5), picked(c_6_1), picked(c_4_9), picked(c_4_5)
    picked(c_7_5), picked(c_0_1), picked(c_3_8)
\end{lstlisting}
\lstset{
    basicstyle=\footnotesize\ttfamily,
    keywordstyle=\bfseries, 
    morekeywords=[1]{Domain},
    keywordstyle=[1]\color{red}\bfseries,
    morekeywords=[2]{Objects, Primitive, Goal, Initial, Plan, plan},
    keywordstyle=[2]\bfseries,
    morekeywords=[3]{pick, reward, move },
    keywordstyle=[3]\color{orange}\bfseries,
    morekeywords=[4]{c\_0\_0, c\_0\_1, c\_0\_2, c\_0\_3, c\_0\_4, c\_0\_5, c\_0\_6, c\_0\_7, c\_0\_8, c\_0\_9, c\_0\_10, c\_0\_11, c\_0\_12, c\_0\_13, c\_0\_14, c\_0\_15, c\_0\_16,
c\_1\_0, c\_1\_1, c\_1\_2, c\_1\_3, c\_1\_4, c\_1\_5, c\_1\_6, c\_1\_7, c\_1\_8, c\_1\_9, c\_1\_10, c\_1\_11, c\_1\_12, c\_1\_13, c\_1\_14, c\_1\_15, c\_1\_16,
c\_2\_0, c\_2\_1, c\_2\_2, c\_2\_3, c\_2\_4, c\_2\_5, c\_2\_6, c\_2\_7, c\_2\_8, c\_2\_9, c\_2\_10, c\_2\_11, c\_2\_12, c\_2\_13, c\_2\_14, c\_2\_15, c\_2\_16,
c\_3\_0, c\_3\_1, c\_3\_2, c\_3\_3, c\_3\_4, c\_3\_5, c\_3\_6, c\_3\_7, c\_3\_8, c\_3\_9, c\_3\_10, c\_3\_11, c\_3\_12, c\_3\_13, c\_3\_14, c\_3\_15, c\_3\_16,
c\_4\_0, c\_4\_1, c\_4\_2, c\_4\_3, c\_4\_4, c\_4\_5, c\_4\_6, c\_4\_7, c\_4\_8, c\_4\_9, c\_4\_10, c\_4\_11, c\_4\_12, c\_4\_13, c\_4\_14, c\_4\_15, c\_4\_16,
c\_5\_0, c\_5\_1, c\_5\_2, c\_5\_3, c\_5\_4, c\_5\_5, c\_5\_6, c\_5\_7, c\_5\_8, c\_5\_9, c\_5\_10, c\_5\_11, c\_5\_12, c\_5\_13, c\_5\_14, c\_5\_15, c\_5\_16,
c\_6\_0, c\_6\_1, c\_6\_2, c\_6\_3, c\_6\_4, c\_6\_5, c\_6\_6, c\_6\_7, c\_6\_8, c\_6\_9, c\_6\_10, c\_6\_11, c\_6\_12, c\_6\_13, c\_6\_14, c\_6\_15, c\_6\_16,
c\_7\_0, c\_7\_1, c\_7\_2, c\_7\_3, c\_7\_4, c\_7\_5, c\_7\_6, c\_7\_7, c\_7\_8, c\_7\_9, c\_7\_10, c\_7\_11, c\_7\_12, c\_7\_13, c\_7\_14, c\_7\_15, c\_7\_16,
c\_8\_0, c\_8\_1, c\_8\_2, c\_8\_3, c\_8\_4, c\_8\_5, c\_8\_6, c\_8\_7, c\_8\_8, c\_8\_9, c\_8\_10, c\_8\_11, c\_8\_12, c\_8\_13, c\_8\_14, c\_8\_15, c\_8\_16,
c\_9\_0, c\_9\_1, c\_9\_2, c\_9\_3, c\_9\_4, c\_9\_5, c\_9\_6, c\_9\_7, c\_9\_8, c\_9\_9, c\_9\_10, c\_9\_11, c\_9\_12, c\_9\_13, c\_9\_14, c\_9\_15, c\_9\_16,
c\_10\_0, c\_10\_1, c\_10\_2, c\_10\_3, c\_10\_4, c\_10\_5, c\_10\_6, c\_10\_7, c\_10\_8, c\_10\_9, c\_10\_10, c\_10\_11, c\_10\_12, c\_10\_13, c\_10\_14, c\_10\_15, c\_10\_16,
c\_11\_0, c\_11\_1, c\_11\_2, c\_11\_3, c\_11\_4, c\_11\_5, c\_11\_6, c\_11\_7, c\_11\_8, c\_11\_9, c\_11\_10, c\_11\_11, c\_11\_12, c\_11\_13, c\_11\_14, c\_11\_15, c\_11\_16,
c\_12\_0, c\_12\_1, c\_12\_2, c\_12\_3, c\_12\_4, c\_12\_5, c\_12\_6, c\_12\_7, c\_12\_8, c\_12\_9, c\_12\_10, c\_12\_11, c\_12\_12, c\_12\_13, c\_12\_14, c\_12\_15, c\_12\_16,
c\_13\_0, c\_13\_1, c\_13\_2, c\_13\_3, c\_13\_4, c\_13\_5, c\_13\_6, c\_13\_7, c\_13\_8, c\_13\_9, c\_13\_10, c\_13\_11, c\_13\_12, c\_13\_13, c\_13\_14, c\_13\_15, c\_13\_16,
c\_14\_0, c\_14\_1, c\_14\_2, c\_14\_3, c\_14\_4, c\_14\_5, c\_14\_6, c\_14\_7, c\_14\_8, c\_14\_9, c\_14\_10, c\_14\_11, c\_14\_12, c\_14\_13, c\_14_14, c\_14\_15, c\_14\_16,
c\_15\_0, c\_15\_1, c\_15\_2, c\_15\_3, c\_15\_4, c\_15\_5, c\_15\_6, c\_15\_7, c\_15\_8, c\_15\_9, c\_15\_10, c\_15\_11, c\_15\_12, c\_15\_13, c\_15\_14, c\_15\_15, c\_15\_16,
c\_16\_0, c\_16\_1, c\_16\_2, c\_16\_3, c\_16\_4, c\_16\_5, c\_16\_6, c\_16\_7, c\_16\_8, c\_16\_9, c\_16\_10, c\_16\_11, c\_16\_12, c\_16\_13, c\_16\_14, c\_16\_15, c\_16\_16},
    keywordstyle=[4]\color{blue}\bfseries,
}

\begin{lstlisting}
Primitive plan: 53
Plan: 8
1 move(c_0_0, c_0_1) -> move(c_0_1, c_1_1) -> move(c_1_1, c_2_1)
   ->move(c_2_1, c_3_1) -> move(c_3_1, c_4_1) -> move(c_4_1, c_5_1)
   ->move(c_5_1, c_6_1) -> pick-reward(c_6_1)
2 move(c_6_1, c_5_1) -> move(c_5_1, c_5_2) -> move(c_5_2, c_4_2)
   ->move(c_4_2, c_4_3) -> move(c_4_3, c_4_4) -> move(c_4_4, c_4_5)
   ->pick-reward(c_4_5)
3 move(c_4_5, c_3_5) -> move(c_3_5, c_3_4) -> move(c_3_4, c_2_4)
   ->move(c_2_4, c_2_3) -> move(c_2_3, c_1_3) -> move(c_1_3, c_1_2)
   ->move(c_1_2, c_1_1) -> move(c_1_1, c_0_1) -> pick-reward(c_0_1)
4 move(c_0_1, c_0_2) -> move(c_0_2, c_0_3) -> move(c_0_3, c_0_4)
   ->move(c_0_4, c_0_5) -> pick-reward(c_0_5)
5 move(c_0_5, c_1_5) -> move(c_1_5, c_2_5) -> move(c_2_5, c_3_5)
   ->move(c_3_5, c_4_5) -> move(c_4_5, c_5_5) -> move(c_5_5, c_6_5)
   ->move(c_6_5, c_7_5) -> move(c_7_5, c_8_5) -> pick-reward(c_8_5)
6 move(c_8_5, c_7_5) -> pick-reward(c_7_5)
7 move(c_7_5, c_7_6) -> move(c_7_6, c_7_7) -> move(c_7_7, c_6_7)
   ->move(c_6_7, c_6_8) -> move(c_6_8, c_5_8) -> move(c_5_8, c_5_9)
   ->move(c_5_9, c_4_9) -> move(c_4_9, c_3_9) -> move(c_3_9, c_3_8)
   ->pick-reward(c_3_8)
8 move(c_3_8, c_3_9) -> move(c_3_9, c_4_9) -> pick-reward(c_4_9)
\end{lstlisting}

\listingcaption{Generated \siwone\ plan for Reward with $10 \times 10$ grid size and $8$ rewards.}
\label{lst:reward_1}
\end{figure*}

\begin{figure*}
\footnotesize
\begin{lstlisting}[basicstyle=\footnotesize\ttfamily]
Name: reward-12x12 (instance_12x12_0.pddl)
Objects: 
    c_0_0, c_0_1, ..., c_11_11, 
Initial: 
    at(c_0_0), reward(c_0_4), reward(c_10_8), reward(c_11_9), reward(c_1_4)
    reward(c_2_8), reward(c_3_8), reward(c_3_9), reward(c_4_6), reward(c_6_6)
    reward(c_9_11), unblocked(c_0_0), ..., unblocked(c_11_11), 
    adjacent(c_0_1, c_0_0), adjacent(c_0_0, c_0_1), ..., adjacent(c_11_10, c_11_11)
Goal: 
    picked(c_10_8), picked(c_11_9), picked(c_3_9), picked(c_1_4), picked(c_4_6)
    picked(c_0_4), picked(c_2_8), picked(c_6_6), picked(c_9_11), picked(c_3_8)
\end{lstlisting}
\lstset{
    basicstyle=\footnotesize\ttfamily,
    keywordstyle=\bfseries, 
    morekeywords=[1]{Domain},
    keywordstyle=[1]\color{red}\bfseries,
    morekeywords=[2]{Objects, Primitive, Goal, Initial, Plan, plan},
    keywordstyle=[2]\bfseries,
    morekeywords=[3]{pick, reward, move },
    keywordstyle=[3]\color{orange}\bfseries,
    morekeywords=[4]{c\_0\_0, c\_0\_1, c\_0\_2, c\_0\_3, c\_0\_4, c\_0\_5, c\_0\_6, c\_0\_7, c\_0\_8, c\_0\_9, c\_0\_10, c\_0\_11, c\_0\_12, c\_0\_13, c\_0\_14, c\_0\_15, c\_0\_16,
c\_1\_0, c\_1\_1, c\_1\_2, c\_1\_3, c\_1\_4, c\_1\_5, c\_1\_6, c\_1\_7, c\_1\_8, c\_1\_9, c\_1\_10, c\_1\_11, c\_1\_12, c\_1\_13, c\_1\_14, c\_1\_15, c\_1\_16,
c\_2\_0, c\_2\_1, c\_2\_2, c\_2\_3, c\_2\_4, c\_2\_5, c\_2\_6, c\_2\_7, c\_2\_8, c\_2\_9, c\_2\_10, c\_2\_11, c\_2\_12, c\_2\_13, c\_2\_14, c\_2\_15, c\_2\_16,
c\_3\_0, c\_3\_1, c\_3\_2, c\_3\_3, c\_3\_4, c\_3\_5, c\_3\_6, c\_3\_7, c\_3\_8, c\_3\_9, c\_3\_10, c\_3\_11, c\_3\_12, c\_3\_13, c\_3\_14, c\_3\_15, c\_3\_16,
c\_4\_0, c\_4\_1, c\_4\_2, c\_4\_3, c\_4\_4, c\_4\_5, c\_4\_6, c\_4\_7, c\_4\_8, c\_4\_9, c\_4\_10, c\_4\_11, c\_4\_12, c\_4\_13, c\_4\_14, c\_4\_15, c\_4\_16,
c\_5\_0, c\_5\_1, c\_5\_2, c\_5\_3, c\_5\_4, c\_5\_5, c\_5\_6, c\_5\_7, c\_5\_8, c\_5\_9, c\_5\_10, c\_5\_11, c\_5\_12, c\_5\_13, c\_5\_14, c\_5\_15, c\_5\_16,
c\_6\_0, c\_6\_1, c\_6\_2, c\_6\_3, c\_6\_4, c\_6\_5, c\_6\_6, c\_6\_7, c\_6\_8, c\_6\_9, c\_6\_10, c\_6\_11, c\_6\_12, c\_6\_13, c\_6\_14, c\_6\_15, c\_6\_16,
c\_7\_0, c\_7\_1, c\_7\_2, c\_7\_3, c\_7\_4, c\_7\_5, c\_7\_6, c\_7\_7, c\_7\_8, c\_7\_9, c\_7\_10, c\_7\_11, c\_7\_12, c\_7\_13, c\_7\_14, c\_7\_15, c\_7\_16,
c\_8\_0, c\_8\_1, c\_8\_2, c\_8\_3, c\_8\_4, c\_8\_5, c\_8\_6, c\_8\_7, c\_8\_8, c\_8\_9, c\_8\_10, c\_8\_11, c\_8\_12, c\_8\_13, c\_8\_14, c\_8\_15, c\_8\_16,
c\_9\_0, c\_9\_1, c\_9\_2, c\_9\_3, c\_9\_4, c\_9\_5, c\_9\_6, c\_9\_7, c\_9\_8, c\_9\_9, c\_9\_10, c\_9\_11, c\_9\_12, c\_9\_13, c\_9\_14, c\_9\_15, c\_9\_16,
c\_10\_0, c\_10\_1, c\_10\_2, c\_10\_3, c\_10\_4, c\_10\_5, c\_10\_6, c\_10\_7, c\_10\_8, c\_10\_9, c\_10\_10, c\_10\_11, c\_10\_12, c\_10\_13, c\_10\_14, c\_10\_15, c\_10\_16,
c\_11\_0, c\_11\_1, c\_11\_2, c\_11\_3, c\_11\_4, c\_11\_5, c\_11\_6, c\_11\_7, c\_11\_8, c\_11\_9, c\_11\_10, c\_11\_11, c\_11\_12, c\_11\_13, c\_11\_14, c\_11\_15, c\_11\_16,
c\_12\_0, c\_12\_1, c\_12\_2, c\_12\_3, c\_12\_4, c\_12\_5, c\_12\_6, c\_12\_7, c\_12\_8, c\_12\_9, c\_12\_10, c\_12\_11, c\_12\_12, c\_12\_13, c\_12\_14, c\_12\_15, c\_12\_16,
c\_13\_0, c\_13\_1, c\_13\_2, c\_13\_3, c\_13\_4, c\_13\_5, c\_13\_6, c\_13\_7, c\_13\_8, c\_13\_9, c\_13\_10, c\_13\_11, c\_13\_12, c\_13\_13, c\_13\_14, c\_13\_15, c\_13\_16,
c\_14\_0, c\_14\_1, c\_14\_2, c\_14\_3, c\_14\_4, c\_14\_5, c\_14\_6, c\_14\_7, c\_14\_8, c\_14\_9, c\_14\_10, c\_14\_11, c\_14\_12, c\_14\_13, c\_14_14, c\_14\_15, c\_14\_16,
c\_15\_0, c\_15\_1, c\_15\_2, c\_15\_3, c\_15\_4, c\_15\_5, c\_15\_6, c\_15\_7, c\_15\_8, c\_15\_9, c\_15\_10, c\_15\_11, c\_15\_12, c\_15\_13, c\_15\_14, c\_15\_15, c\_15\_16,
c\_16\_0, c\_16\_1, c\_16\_2, c\_16\_3, c\_16\_4, c\_16\_5, c\_16\_6, c\_16\_7, c\_16\_8, c\_16\_9, c\_16\_10, c\_16\_11, c\_16\_12, c\_16\_13, c\_16\_14, c\_16\_15, c\_16\_16},
    keywordstyle=[4]\color{blue}\bfseries,
}

\begin{lstlisting}
Primitive plan: 74
Plan: 10
1  move(c_0_0, c_0_1) -> move(c_0_1, c_1_1) -> move(c_1_1, c_1_2)
    ->move(c_1_2, c_2_2) -> move(c_2_2, c_3_2) -> move(c_3_2, c_4_2)
    ->move(c_4_2, c_5_2) -> move(c_5_2, c_6_2) -> move(c_6_2, c_6_3)
    ->move(c_6_3, c_7_3) -> move(c_7_3, c_8_3) -> move(c_8_3, c_8_4)
    ->move(c_8_4, c_9_4) -> move(c_9_4, c_9_5) -> move(c_9_5, c_9_6)
    ->move(c_9_6, c_9_7) -> move(c_9_7, c_9_8) -> move(c_9_8, c_9_9)
    ->move(c_9_9, c_9_10) -> move(c_9_10, c_9_11) -> pick-reward(c_9_11)
2  move(c_9_11, c_9_10) -> move(c_9_10, c_8_10) -> move(c_8_10, c_8_9)
    ->move(c_8_9, c_8_8) -> move(c_8_8, c_8_7) -> move(c_8_7, c_8_6)
    ->move(c_8_6, c_7_6) -> move(c_7_6, c_6_6) -> pick-reward(c_6_6)
3  move(c_6_6, c_5_6) -> move(c_5_6, c_4_6) -> pick-reward(c_4_6)
4  move(c_4_6, c_4_5) -> move(c_4_5, c_4_4) -> move(c_4_4, c_3_4)
    ->move(c_3_4, c_2_4) -> move(c_2_4, c_1_4) -> pick-reward(c_1_4)
5  move(c_1_4, c_0_4) -> pick-reward(c_0_4)
6  move(c_0_4, c_1_4) -> move(c_1_4, c_2_4) -> move(c_2_4, c_2_5)
    ->move(c_2_5, c_2_6) -> move(c_2_6, c_3_6) -> move(c_3_6, c_4_6)
    ->move(c_4_6, c_5_6) -> move(c_5_6, c_6_6) -> move(c_6_6, c_7_6)
    ->move(c_7_6, c_8_6) -> move(c_8_6, c_8_7) -> move(c_8_7, c_9_7)
    ->move(c_9_7, c_9_8) -> move(c_9_8, c_10_8) -> pick-reward(c_10_8)
7  move(c_10_8, c_10_9) -> move(c_10_9, c_11_9) -> pick-reward(c_11_9)
8  move(c_11_9, c_10_9) -> move(c_10_9, c_9_9) -> move(c_9_9, c_8_9)
    ->move(c_8_9, c_7_9) -> move(c_7_9, c_6_9) -> move(c_6_9, c_5_9)
    ->move(c_5_9, c_4_9) -> move(c_4_9, c_3_9) -> move(c_3_9, c_3_8)
    ->pick-reward(c_3_8)
9  move(c_3_8, c_2_8) -> pick-reward(c_2_8)
10 move(c_2_8, c_3_8) -> move(c_3_8, c_3_9) -> pick-reward(c_3_9)
\end{lstlisting}
\listingcaption{Generated \siwone\ plan for Reward with $12 \times 12$ grid size and $10$ rewards.}
\label{lst:reward_2}
\end{figure*}

\lstset{
    basicstyle=\footnotesize\ttfamily,
    keywordstyle=\bfseries, 
    morekeywords=[1]{Domain},
    keywordstyle=[1]\color{red}\bfseries,
    morekeywords=[2]{Objects, Primitive, Goal, Initial, Plan, plan},
    keywordstyle=[2]\bfseries,
    morekeywords=[3]{walk, tighten_nut, pickup_spanner },
    keywordstyle=[3]\color{orange}\bfseries,
    morekeywords=[4]{spanner21, shed, spanner14, spanner20, spanner24, spanner13, nut5, spanner4, spanner16, nut1, spanner2, spanner6, nut3, spanner11, gate, nut11, location6, location7, spanner22, spanner7, spanner17,      nut2, spanner8, spanner3, nut6, nut8, location8, nut9, location2, nut7, nut4, location1, spanner5, spanner19, bob, location4, spanner12, location5, spanner9, spanner15, location9, location10, spanner10, nut10, location3, spanner18, nut12, spanner23, spanner1,
    spanner25, spanner26, spanner27, spanner28, spanner29, spanner30, spanner31, spanner32, spanner33, spanner34, spanner35, spanner36, spanner37, spanner38, spanner39, spanner40,
    nut13, nut14, nut15, nut16, 
    location11, location12, location13, location14, location15, location16, location17, location18, location19, location20},
    keywordstyle=[4]\color{blue}\bfseries,
}

\begin{figure*}
\footnotesize
\begin{lstlisting}[basicstyle=\footnotesize\ttfamily]
Domain: spanner
Types: locatable, location, man, nut, object, spanner
Predicates: at/2, carrying/2, link/2, loose/1, tightened/1, useable/1
Action schemas:
walk(?start: location, ?end: location, ?m: man)
pre:  +at(?m, ?start),  +link(?start, ?end)
eff:  -at(?m, ?start),  +at(?m, ?end)
pickup_spanner(?l: location, ?s: spanner, ?m: man)
pre:  +at(?m, ?l),  +at(?s, ?l)
eff:  -at(?s, ?l),  +carrying(?m, ?s)
tighten_nut(?l: location, ?s: spanner, ?m: man, ?n: nut)
pre:  +at(?m, ?l),  +at(?n, ?l),  +carrying(?m, ?s),  +useable(?s),  +loose(?n)
eff:  -loose(?n),  -useable(?s),  +tightened(?n)

Name: prob (spanner_s-10_n-10_l-10.pddl)
Objects: 
    bob, gate, location1, ..., location10, nut1, ..., nut10, shed, spanner1, ..., spanner10
Initial: 
    at(spanner1, location1), at(spanner9, location3), at(spanner10, location6)
    at(spanner3, location7), at(spanner4, location7), at(spanner7, location7)
    at(spanner2, location8), at(spanner5, location8), at(spanner6, location8)
    at(spanner8, location10), at(bob, shed), at(nut1, gate), ..., at(nut10, gate),
    useable(spanner1), ..., useable(spanner10)
    link(shed, location1), link(location1, location2), ..., 
    link(location9, location10), link(location10, gate)
    loose(nut1), ..., loose(nut10)
Goal: 
    tightened(nut1), ..., tightened(nut10)

Primitive plan: 31
Plan: 20
1  walk(shed, location1, bob) -> pickup_spanner(location1, spanner1, bob)
2  walk(location1, location2, bob) -> walk(location2, location3, bob) 
    ->pickup_spanner(location3, spanner9, bob)
3  walk(location3, location4, bob) -> walk(location4, location5, bob)
    ->walk(location5, location6, bob) -> pickup_spanner(location6, spanner10, bob)
4  walk(location6, location7, bob) -> pickup_spanner(location7, spanner4, bob)
5  pickup_spanner(location7, spanner3, bob)
6  pickup_spanner(location7, spanner7, bob)
7  walk(location7, location8, bob) -> pickup_spanner(location8, spanner5, bob)
8  pickup_spanner(location8, spanner6, bob)
9  pickup_spanner(location8, spanner2, bob)
10 walk(location8, location9, bob) -> walk(location9, location10, bob) 
    ->pickup_spanner(location10, spanner8, bob)
11 walk(location10, gate, bob) -> tighten_nut(gate, spanner7, bob, nut10)
12 tighten_nut(gate, spanner8, bob, nut2)
13 tighten_nut(gate, spanner4, bob, nut4)
14 tighten_nut(gate, spanner3, bob, nut7)
15 tighten_nut(gate, spanner10, bob, nut8)
16 tighten_nut(gate, spanner1, bob, nut9)
17 tighten_nut(gate, spanner6, bob, nut1)
18 tighten_nut(gate, spanner9, bob, nut3)
19 tighten_nut(gate, spanner5, bob, nut6)
20 tighten_nut(gate, spanner2, bob, nut5)
\end{lstlisting}
\listingcaption{Generated \siwone\ plan for Spanner with $10$ spanners, $10$ nuts, and $10$ locations.}
\label{lst:spanner_1}
\end{figure*}

\begin{figure*}
\footnotesize
\begin{lstlisting}[basicstyle=\footnotesize\ttfamily]
Name: prob (spanner_s-14_n-7_l-10.pddl)
Objects: 
    bob, gate, location, ..., location10, nut1, ..., nut7, shed, 
    spanner1, ..., spanner14, 
Initial: 
    at(spanner5, location2), at(spanner8, location3), at(spanner1, location4)
    at(spanner9, location4), at(spanner4, location5), at(spanner10, location5)
    at(spanner14, location5), at(spanner3, location6), at(spanner7, location7)
    at(spanner6, location8), at(spanner12, location8), at(spanner13, location8)
    at(spanner2, location9), at(spanner11, location10), at(bob, shed), at(nut1
    gate), at(nut2, gate), at(nut3, gate), at(nut4, gate), at(nut5, gate), at(nut6
    gate), at(nut7, gate), useable(spanner1), ..., useable(spanner14), link(shed
    location1), link(location1, location2), ..., link(location9, location10),
    link(location10, gate), loose(nut1), ..., loose(nut7)
Goal: 
    tightened(nut1), ..., tightened(nut7)

Primitive plan: 31
Plan: 21
1  walk(shed, location1, bob) -> walk(location1, location2, bob)
2  pickup_spanner(location2, spanner5, bob)
3  walk(location2, location3, bob) -> pickup_spanner(location3, spanner8, bob)
4  walk(location3, location4, bob) -> pickup_spanner(location4, spanner1, bob)
5  pickup_spanner(location4, spanner9, bob)
6  walk(location4, location5, bob) -> pickup_spanner(location5, spanner4, bob)
7  pickup_spanner(location5, spanner10, bob)
8  pickup_spanner(location5, spanner14, bob)
9  walk(location5, location6, bob) -> pickup_spanner(location6, spanner3, bob)
10 walk(location6, location7, bob) -> pickup_spanner(location7, spanner7, bob)
11 walk(location7, location8, bob) -> pickup_spanner(location8, spanner6, bob)
12 pickup_spanner(location8, spanner12, bob)
13 pickup_spanner(location8, spanner13, bob)
14 walk(location8, location9, bob) -> pickup_spanner(location9, spanner2, bob)
15 walk(location9, location10, bob) -> walk(location10, gate, bob) 
    ->tighten_nut(gate, spanner13, bob, nut5)
16 tighten_nut(gate, spanner5, bob, nut3)
17 tighten_nut(gate, spanner2, bob, nut1)
18 tighten_nut(gate, spanner9, bob, nut4)
19 tighten_nut(gate, spanner3, bob, nut7)
20 tighten_nut(gate, spanner10, bob, nut6)
21 tighten_nut(gate, spanner8, bob, nut2
\end{lstlisting}
\listingcaption{Generated \siwone\ plan for Spanner with $14$ spanners, $7$ nuts, and $10$ locations.}
\label{lst:spanner_2}
\end{figure*}

\begin{figure*}
\footnotesize
\begin{lstlisting}[basicstyle=\footnotesize\ttfamily]
Name: prob (spanner_s-16_n-8_l-20.pddl)
Objects: 
    bob, gate, location1, ..., location20, nut1, ..., nut8, shed, spanner1, ..., spanner16
Initial: 
    at(spanner2, location1), at(spanner14, location4), at(spanner5, location5)
    at(spanner6, location5), at(spanner13, location5), at(spanner7, location6)
    at(spanner4, location8), at(spanner1, location10), at(spanner15, location11)
    at(spanner9, location13), at(spanner10, location13), at(spanner8, location14)
    at(spanner16, location15), at(spanner12, location18), at(spanner3, location20)
    at(spanner11, location20), at(bob, shed), at(nut1, gate), ..., at(nut5, gate),
    useable(spanner1), ..., useable(spanner16)
    link(shed, location1), link(location1, location2), ..., 
    link(location19, location20), link(location20, gate), loose(nut1), ..., loose(nut8)
Goal: 
    tightened(nut1), ..., tightened(nut8)

Primitive plan: 44
Plan: 23
1  walk(shed, location1, bob) -> pickup_spanner(location1, spanner2, bob)
2  walk(location1, location2, bob) -> walk(location2, location3, bob) 
    -> walk(location3, location4, bob) -> pickup_spanner(location4, spanner14, bob)
3  walk(location4, location5, bob) -> pickup_spanner(location5, spanner6, bob)
4  pickup_spanner(location5, spanner5, bob)
5  pickup_spanner(location5, spanner13, bob)
6  walk(location5, location6, bob) -> pickup_spanner(location6, spanner7, bob)
7  walk(location6, location7, bob) -> walk(location7, location8, bob) 
    ->pickup_spanner(location8, spanner4, bob)
8  walk(location8, location9, bob) -> walk(location9, location10, bob) 
    ->pickup_spanner(location10, spanner1, bob)
9  walk(location10, location11, bob) -> pickup_spanner(location11, spanner15, bob)
10 walk(location11, location12, bob) -> walk(location12, location13, bob) 
    ->pickup_spanner(location13, spanner9, bob)
11 pickup_spanner(location13, spanner10, bob)
12 walk(location13, location14, bob) -> pickup_spanner(location14, spanner8, bob)
13 walk(location14, location15, bob) -> pickup_spanner(location15, spanner16, bob)
14 walk(location15, location16, bob) -> walk(location16, location17, bob)
    ->walk(location17, location18, bob) -> pickup_spanner(location18, spanner12, bob)
15 walk(location18, location19, bob) -> walk(location19, location20, bob) 
    -> pickup_spanner(location20, spanner3, bob)
16 walk(location20, gate, bob) -> tighten_nut(gate, spanner16, bob, nut2)
17 tighten_nut(gate, spanner9, bob, nut8)
18 tighten_nut(gate, spanner12, bob, nut3)
19 tighten_nut(gate, spanner7, bob, nut5)
20 tighten_nut(gate, spanner4, bob, nut1)
21 tighten_nut(gate, spanner3, bob, nut7)
22 tighten_nut(gate, spanner15, bob, nut6)
23 tighten_nut(gate, spanner10, bob, nut4
\end{lstlisting}
\listingcaption{Generated \siwone\ plan for Spanner with $16$ spanners, $8$ nuts, and $20$ locations.}
\label{lst:spanner_3}
\end{figure*}

\lstset{
    basicstyle=\footnotesize\ttfamily,
    keywordstyle=\bfseries, 
    morekeywords=[1]{Domain},
    keywordstyle=[1]\color{red}\bfseries,
    morekeywords=[2]{Objects, Primitive, Goal, Initial, Plan, plan},
    keywordstyle=[2]\bfseries,
    morekeywords=[3]{pick, up, put, down, stack, unstack },
    keywordstyle=[3]\color{orange}\bfseries,
    morekeywords=[4]{a,b,c,d,e,f,g,h,i,j,k,l,m,n,o,p,q,r,s,t,u,v,w,b1,b2,b3,b4,b5,b6,b7,b8,b9,b10,b11,b12,b13,b14,b15,b16,b17,b18,b19,b20},
    keywordstyle=[4]\color{blue}\bfseries,
}

\begin{figure*}
\footnotesize
\begin{lstlisting}[basicstyle=\footnotesize\ttfamily]
Domain: blocks
Types: object
Predicates: clear/1, handempty/0, holding/1, on/2, ontable/1
Action schemas:
pick-up(?x: object)
pre:  +clear(?x),  +ontable(?x),  +handempty()
eff:  -ontable(?x),  -clear(?x),  -handempty(),  +holding(?x)
put-down(?x: object)
pre:  +holding(?x)
eff:  -holding(?x),  +clear(?x),  +handempty(),  +ontable(?x)
stack(?x: object, ?y: object)
pre:  +holding(?x),  +clear(?y)
eff:  -holding(?x),  -clear(?y),  +clear(?x),  +handempty(),  +on(?x, ?y)
unstack(?x: object, ?y: object)
pre:  +on(?x, ?y),  +clear(?x),  +handempty()
eff:  +holding(?x),  +clear(?y),  -clear(?x),  -handempty(),  -on(?x, ?y)

Name: blocks-10-0 (probBLOCKS-10-0.pddl)
Objects: 
    a, b, c, d, e, f, g, h, i, j
Initial: 
    on(a, d), on(h, a), on(g, h), on(b, g), on(j, b), on(e, j), on(c, e), on(d, i)
    ontable(i), ontable(f), clear(f), clear(c), handempty()
Goal: 
    on(d, c), on(c, f), on(f, j), on(j, e), on(e, h), on(h, b), on(b, a), on(a, g)
    on(g, i)

Primitive plan: 46
Plan: 12
1  unstack(c, e) -> stack(c, f) -> unstack(e, j)
    ->stack(e, c) -> unstack(j, b) -> stack(j, e)
    ->unstack(b, g) -> stack(b, j) -> unstack(g, h)
    ->stack(g, b) -> unstack(h, a) -> stack(h, g)
    ->unstack(a, d) -> stack(a, h) -> unstack(d, i)
    ->put-down(d)
2  unstack(a, h) -> stack(a, d) -> unstack(h, g)
    ->stack(h, a) -> unstack(g, b) -> stack(g, i)
3  unstack(h, a) -> stack(h, b)
4  unstack(a, d) -> stack(a, g)
5  unstack(h, b) -> put-down(h)
6  unstack(b, j) -> stack(b, a)
7  pick-up(h) -> stack(h, b)
8  unstack(j, e) -> put-down(j)
9  unstack(e, c) -> stack(e, h) -> unstack(c, f)
    ->stack(c, d)
10 pick-up(j) -> stack(j, e)
11 pick-up(f) -> stack(f, j)
12 unstack(c, d) -> stack(c, f) -> pick-up(d)
    ->stack(d, c)
\end{lstlisting}
\listingcaption{Generated \siwone\ plan for Blocksworld with single tower goal and 10 blocks.}
\label{lst:blocks_single_1}
\end{figure*}

\begin{figure*}
\begin{lstlisting}
Name: blocks-10-0 (probBLOCKS-10-0.pddl)
\end{lstlisting}
See listing \ref{lst:blocks_single_1} for problem specifics.
\begin{lstlisting}
Primitive plan: 40
Plan: 5
1 unstack(c, e) -> stack(c, f) -> unstack(e, j)
   ->stack(e, c) -> unstack(j, b) -> stack(j, e)
   ->unstack(b, g) -> stack(b, j) -> unstack(g, h)
   ->stack(g, b) -> unstack(h, a) -> put-down(h)
   ->unstack(a, d) -> stack(a, h) -> unstack(d, i)
   ->put-down(d) -> unstack(g, b) -> stack(g, i)
   ->unstack(a, h) -> stack(a, g) -> pick-up(h)
2 stack(h, d) -> unstack(b, j) -> stack(b, a)
   ->unstack(h, d) -> stack(h, b)
3 unstack(j, e) -> stack(j, d) -> unstack(e, c)
   ->stack(e, h) -> unstack(j, d) -> stack(j, e)
4 unstack(c, f) -> stack(c, d) -> pick-up(f)
   ->stack(f, j) -> unstack(c, d) -> stack(c, f)
5 pick-up(d) -> stack(d, c)
\end{lstlisting}
\listingcaption{Generated \siwtwo\ plan for Blocksworld with single tower goal and 10 blocks.}
\label{lst:blocks_single_3}
\end{figure*}

\begin{figure*}
\footnotesize
\begin{lstlisting}[basicstyle=\footnotesize\ttfamily]
Name: blocks-20-0 (probBLOCKS-20-0.pddl)
Objects: 
    b1, ..., b20
Initial: 
    on(b14, b1), on(b2, b10), on(b19, b11), on(b17, b12), on(b4, b13), on(b6, b14)
    on(b5, b15), on(b10, b16), on(b13, b17), on(b9, b18), on(b15, b19), on(b18, b2)
    on(b3, b20), on(b1, b3), on(b11, b4), on(b16, b6), on(b20, b7), on(b12, b8)
    on(b8, b9), ontable(b7), clear(b5), handempty()
Goal: 
    on(b1, b20), on(b10, b17), on(b11, b5), on(b12, b4), on(b13, b1), on(b14, b6)
    on(b15, b19), on(b16, b15), on(b17, b8), on(b18, b14), on(b19, b7), on(b2, b9)
    on(b20, b18), on(b3, b11), on(b5, b13), on(b6, b10), on(b7, b2), on(b8, b16)
    on(b9, b12)
    
Primitive plan: 88
Plan: 30
1  unstack(b5, b15) -> put-down(b5) -> unstack(b15, b19)
    ->stack(b15, b5) -> unstack(b19, b11) -> stack(b19, b15)
    ->unstack(b11, b4) -> stack(b11, b19) -> unstack(b4, b13)
    ->put-down(b4)
2  unstack(b13, b17) -> stack(b13, b11) -> unstack(b17, b12)
    ->stack(b17, b13) -> unstack(b12, b8) -> stack(b12, b4)
3  unstack(b8, b9) -> stack(b8, b17) -> unstack(b9, b18)
    ->stack(b9, b12) -> unstack(b18, b2) -> put-down(b18)
4  unstack(b2, b10) -> stack(b2, b9)
5  unstack(b8, b17) -> put-down(b8)
6  unstack(b17, b13) -> stack(b17, b8)
7  unstack(b13, b11) -> put-down(b13)
8  unstack(b11, b19) -> stack(b11, b13) -> unstack(b19, b15)
    ->put-down(b19)
9  unstack(b15, b5) -> put-down(b15)
10 unstack(b10, b16) -> put-down(b10)
11 unstack(b16, b6) -> put-down(b16)
12 unstack(b6, b14) -> put-down(b6)
13 unstack(b14, b1) -> stack(b14, b6) -> unstack(b1, b3)
    ->put-down(b1)
14 unstack(b3, b20) -> put-down(b3)
15 unstack(b20, b7) -> stack(b20, b1) -> pick-up(b7)
    ->stack(b7, b2)
16 pick-up(b19) -> stack(b19, b7)
17 pick-up(b15) -> stack(b15, b19)
18 pick-up(b16) -> stack(b16, b15)
19 unstack(b11, b13) -> stack(b11, b3)
20 unstack(b17, b8) -> stack(b17, b13) -> pick-up(b8)
    ->stack(b8, b16)
21 unstack(b17, b13) -> stack(b17, b8)
22 pick-up(b10) -> stack(b10, b17)
23 unstack(b14, b6) -> stack(b14, b13) -> pick-up(b6)
    ->stack(b6, b10)
24 unstack(b14, b13) -> stack(b14, b6)
25 pick-up(b18) -> stack(b18, b14)
26 unstack(b20, b1) -> stack(b20, b18)
27 pick-up(b1) -> stack(b1, b20)
28 pick-up(b13) -> stack(b13, b1)
29 pick-up(b5) -> stack(b5, b13)
30 unstack(b11, b3) -> stack(b11, b5) -> pick-up(b3)
    ->stack(b3, b11)
\end{lstlisting}
\listingcaption{Generated \siwone\ plan for Blocksworld with single tower goal and 20 blocks.}
\label{lst:blocks_single_2}
\end{figure*}

\begin{figure*}

\footnotesize
\begin{lstlisting}[basicstyle=\footnotesize\ttfamily]
Name: blocks-20-0 (probBLOCKS-20-0.pddl)
\end{lstlisting}
See listing \ref{lst:blocks_single_2} for problem definition.
\begin{lstlisting}
Primitive plan: 180
Plan: 9
1 unstack(b5, b15) -> put-down(b5) -> unstack(b15, b19)
   ->stack(b15, b5) -> unstack(b19, b11) -> stack(b19, b15)
   ->unstack(b11, b4) -> stack(b11, b19) -> unstack(b4, b13)
   ->stack(b4, b11) -> unstack(b13, b17) -> stack(b13, b4)
   ->unstack(b17, b12) -> stack(b17, b13) -> unstack(b12, b8)
   ->stack(b12, b17) -> unstack(b8, b9) -> stack(b8, b12)
   ->unstack(b9, b18) -> stack(b9, b8) -> unstack(b18, b2)
   ->stack(b18, b9) -> unstack(b2, b10) -> stack(b2, b18)
   ->unstack(b10, b16) -> stack(b10, b2) -> unstack(b16, b6)
   ->stack(b16, b10) -> unstack(b6, b14) -> stack(b6, b16)
   ->unstack(b14, b1) -> stack(b14, b6) -> unstack(b1, b3)
   ->stack(b1, b14) -> unstack(b3, b20) -> put-down(b3)
   ->unstack(b20, b7) -> stack(b20, b3) -> unstack(b1, b14)
   ->stack(b1, b20) -> pick-up(b7) -> stack(b7, b1)
   ->unstack(b14, b6) -> stack(b14, b7) -> unstack(b6, b16)
   ->stack(b6, b14) -> unstack(b16, b10) -> stack(b16, b6)
   ->unstack(b10, b2) -> stack(b10, b16) -> unstack(b2, b18)
   ->stack(b2, b10) -> unstack(b18, b9) -> stack(b18, b2)
   ->unstack(b9, b8) -> stack(b9, b18) -> unstack(b8, b12)
   ->stack(b8, b9) -> unstack(b12, b17) -> stack(b12, b8)
   ->unstack(b17, b13) -> stack(b17, b12) -> unstack(b13, b4)
   ->stack(b13, b17) -> unstack(b4, b11) -> stack(b4, b13)
   ->unstack(b11, b19) -> stack(b11, b4) -> unstack(b19, b15)
   ->stack(b19, b11) -> unstack(b15, b5) -> stack(b15, b19)
   ->pick-up(b5) -> stack(b5, b15)
2 unstack(b5, b15) -> put-down(b5) -> unstack(b15, b19)
   ->stack(b15, b5) -> unstack(b19, b11) -> stack(b19, b15)
   ->unstack(b11, b4) -> stack(b11, b19) -> unstack(b4, b13)
   ->put-down(b4) -> unstack(b13, b17) -> stack(b13, b11)
   ->unstack(b17, b12) -> stack(b17, b13) -> unstack(b12, b8)
   ->stack(b12, b4)
3 unstack(b8, b9) -> stack(b8, b17) -> unstack(b9, b18)
   ->stack(b9, b12) -> unstack(b18, b2) -> stack(b18, b9)
   ->unstack(b2, b10) -> stack(b2, b18) -> unstack(b10, b16)
   ->stack(b10, b8) -> unstack(b16, b6) -> stack(b16, b10)
   ->unstack(b6, b14) -> stack(b6, b16) -> unstack(b14, b7)
   ->stack(b14, b6) -> unstack(b2, b18) -> stack(b2, b7)
   ->unstack(b18, b9) -> stack(b18, b14) -> unstack(b2, b7)
   ->stack(b2, b9)
4 unstack(b7, b1) -> stack(b7, b2) -> unstack(b18, b14)
   ->stack(b18, b1) -> unstack(b14, b6) -> stack(b14, b18)
   ->unstack(b6, b16) -> stack(b6, b14) -> unstack(b16, b10)
   ->stack(b16, b6) -> unstack(b10, b8) -> stack(b10, b16)
   ->unstack(b8, b17) -> stack(b8, b10) -> unstack(b17, b13)
   ->stack(b17, b8) -> unstack(b13, b11) -> stack(b13, b17)
   ->unstack(b11, b19) -> stack(b11, b13) -> unstack(b19, b15)
   ->stack(b19, b7)
5 unstack(b11, b13) -> put-down(b11) -> unstack(b15, b5)
   ->stack(b15, b19) -> unstack(b13, b17) -> stack(b13, b5)
   ->unstack(b17, b8) -> stack(b17, b13) -> unstack(b8, b10)
   ->stack(b8, b17) -> unstack(b10, b16) -> stack(b10, b11)
   ->unstack(b16, b6) -> stack(b16, b15)
6 unstack(b10, b11) -> stack(b10, b6) -> unstack(b8, b17)
   ->stack(b8, b16) -> unstack(b17, b13) -> stack(b17, b8)
   ->pick-up(b11) -> stack(b11, b13) -> unstack(b10, b6)
   ->stack(b10, b17) -> unstack(b6, b14) -> stack(b6, b10)
   ->unstack(b14, b18) -> stack(b14, b6) -> unstack(b18, b1)
   ->stack(b18, b14)
7 unstack(b1, b20) -> stack(b1, b11) -> unstack(b20, b3)
   ->stack(b20, b18) -> unstack(b1, b11) -> stack(b1, b20)
8 unstack(b11, b13) -> stack(b11, b3) -> unstack(b13, b5)
   ->stack(b13, b1) -> pick-up(b5) -> stack(b5, b13)
9 unstack(b11, b3) -> stack(b11, b5) -> pick-up(b3)
   ->stack(b3, b11)
\end{lstlisting}
\listingcaption{Generated \siwtwo\ plan for Blocksworld with single tower goal and 20 blocks.}
\label{lst:blocks_single_4}
\end{figure*}

\lstset{
    basicstyle=\footnotesize\ttfamily,
    keywordstyle=\bfseries, 
    morekeywords=[1]{Domain},
    keywordstyle=[1]\color{red}\bfseries,
    morekeywords=[2]{Objects, Primitive, Goal, Initial, Plan, plan},
    keywordstyle=[2]\bfseries,
    morekeywords=[3]{pick, up, put, down, stack, unstack },
    keywordstyle=[3]\color{orange}\bfseries,
    morekeywords=[4]{b8, b6, b10, b5, b4, b2, b7, b3, b1, b9,b11,b12,b13,b14,b15,b16,b17,b18,b19,20 },
    keywordstyle=[4]\color{blue}\bfseries,
}

\begin{figure*}
\footnotesize
\begin{lstlisting}[basicstyle=\footnotesize\ttfamily]
Domain: blocks
Types: object
Predicates: clear/1, handempty/0, holding/1, on/2, ontable/1
Action schemas:
pick-up(?x: object)
pre:  +clear(?x),  +ontable(?x),  +handempty()
eff:  -ontable(?x),  -clear(?x),  -handempty(),  +holding(?x)
put-down(?x: object)
pre:  +holding(?x)
eff:  -holding(?x),  +clear(?x),  +handempty(),  +ontable(?x)
stack(?x: object, ?y: object)
pre:  +holding(?x),  +clear(?y)
eff:  -holding(?x),  -clear(?y),  +clear(?x),  +handempty(),  +on(?x, ?y)
unstack(?x: object, ?y: object)
pre:  +on(?x, ?y),  +clear(?x),  +handempty()
eff:  +holding(?x),  +clear(?y),  -clear(?x),  -handempty(),  -on(?x, ?y)

Name: blocks-10-0 (probBLOCKS-10-0.pddl)
Objects: 
b1, b10, b2, b3, b4, b5, b6, b7, b8, b9
Initial: 
on(b8, b1), on(b1, b10), on(b10, b3), on(b2, b6), on(b5, b7), on(b9, b8), on(b7
b9), ontable(b3), ontable(b4), ontable(b6), clear(b2), clear(b4), clear(b5)
handempty()
Goal: 
on(b1, b2), on(b10, b5), on(b2, b10), on(b3, b9), on(b4, b8), on(b5, b6), on(b7
b3), on(b8, b7)

Primitive plan: 44
Plan: 15
1  unstack(b5, b7) -> stack(b5, b4) -> unstack(b7, b9)
    ->stack(b7, b2) -> unstack(b9, b8) -> stack(b9, b5)
    ->unstack(b8, b1) -> stack(b8, b9) -> unstack(b1, b10)
    ->put-down(b1)
2  unstack(b8, b9) -> stack(b8, b10) -> unstack(b9, b5)
    ->put-down(b9)
3  unstack(b5, b4)
4  put-down(b5)
5  unstack(b7, b2) -> stack(b7, b5)
6  unstack(b2, b6)
7  stack(b2, b4) -> unstack(b7, b5) -> stack(b7, b9)
    ->pick-up(b5) -> stack(b5, b6)
8  unstack(b8, b10) -> stack(b8, b1) -> unstack(b10, b3)
    ->stack(b10, b5)
9  unstack(b2, b4) -> stack(b2, b10)
10 unstack(b8, b1) -> stack(b8, b4) -> pick-up(b1)
    ->stack(b1, b2)
11 unstack(b7, b9) -> stack(b7, b8)
12 pick-up(b3) -> stack(b3, b9)
13 unstack(b7, b8) -> stack(b7, b3)
14 unstack(b8, b4) -> stack(b8, b7)
15 pick-up(b4) -> stack(b4, b8)
\end{lstlisting}
\listingcaption{Generated \siwone\ plan for Blocksworld with multiple tower goal and 10 blocks.}
\label{lst:blocks_multiple_1}
\end{figure*}

\begin{figure*}
\footnotesize
\begin{lstlisting}[basicstyle=\footnotesize\ttfamily]
Name: blocks-10-0 (probBLOCKS-10-0.pddl)
\end{lstlisting}
See listing \ref{lst:blocks_multiple_1} for problem specifics.
\begin{lstlisting}
Primitive plan: 42
Plan: 5
1 unstack(b2, b6) -> stack(b2, b4) -> unstack(b5, b7)
   ->stack(b5, b6) -> unstack(b7, b9) -> stack(b7, b2)
   ->unstack(b9, b8) -> put-down(b9)
2 unstack(b8, b1) -> stack(b8, b5) -> unstack(b1, b10)
   ->stack(b1, b8) -> unstack(b10, b3) -> stack(b10, b1)
   ->pick-up(b3) -> stack(b3, b9) -> unstack(b7, b2)
   ->stack(b7, b3)
3 unstack(b10, b1) -> stack(b10, b2) -> unstack(b1, b8)
   ->stack(b1, b7) -> unstack(b8, b5) -> stack(b8, b1)
   ->unstack(b10, b2) -> stack(b10, b5) -> unstack(b2, b4)
   ->stack(b2, b10) -> pick-up(b4) -> stack(b4, b8)
4 unstack(b4, b8) -> stack(b4, b2) -> unstack(b8, b1)
   ->stack(b8, b4) -> unstack(b1, b7) -> put-down(b1)
   ->unstack(b8, b4) -> stack(b8, b7) -> unstack(b4, b2)
   ->stack(b4, b8)
5 pick-up(b1) -> stack(b1, b2)
\end{lstlisting}
\listingcaption{Generated \siwtwo\ plan for Blocksworld with multiple tower goal and 10 blocks.}
\label{lst:blocks_multiple_3}
\end{figure*}
            
\begin{figure*}
\footnotesize
\begin{lstlisting}[basicstyle=\footnotesize\ttfamily]
Name: blocks-20-0 (probBLOCKS-20-0.pddl)
Objects: 
    b1, b10, b11, b12, b13, b14, b15, b16, b17, b18, b19, b2, b20, b3, b4, b5, b6
    b7, b8, b9
Initial: 
    on(b13, b10), on(b8, b11), on(b12, b13), on(b15, b14), on(b1, b15), on(b19, b16)
    on(b3, b17), on(b7, b2), on(b5, b20), on(b11, b3), on(b18, b4), on(b10, b5)
    on(b17, b6), on(b6, b7), on(b4, b9), ontable(b14), ontable(b16), ontable(b2)
    ontable(b20), ontable(b9), clear(b1), clear(b12), clear(b18), clear(b19)
    clear(b8), handempty()
Goal: 
    on(b1, b20), on(b10, b16), on(b11, b5), on(b12, b7), on(b13, b12), on(b14, b13)
    on(b15, b10), on(b17, b14), on(b18, b1), on(b19, b15), on(b2, b17), on(b20, b4)
    on(b3, b2), on(b4, b6), on(b5, b8), on(b6, b19), on(b7, b11), on(b9, b3)

Primitive plan: 72
Plan: 19
1  unstack(b18, b4) -> put-down(b18)
2  unstack(b4, b9)
3  stack(b4, b1) -> unstack(b8, b11) -> put-down(b8)
4  unstack(b12, b13) -> stack(b12, b19) -> unstack(b13, b10)
    ->stack(b13, b18) -> unstack(b10, b5) -> stack(b10, b13)
    ->unstack(b5, b20) -> stack(b5, b8)
5  unstack(b11, b3) -> stack(b11, b5) -> unstack(b3, b17)
    ->stack(b3, b20) -> unstack(b17, b6) -> stack(b17, b3)
    ->unstack(b6, b7) -> stack(b6, b17) -> unstack(b7, b2)
    ->stack(b7, b11)
6  unstack(b12, b19) -> stack(b12, b7)
7  unstack(b10, b13) -> stack(b10, b6) -> unstack(b13, b18)
    ->stack(b13, b12)
8  unstack(b19, b16)
9  stack(b19, b13) -> unstack(b10, b6) -> stack(b10, b16)
10 unstack(b4, b1) -> stack(b4, b6) -> unstack(b1, b15)
    ->stack(b1, b2) -> unstack(b15, b14) -> stack(b15, b10)
11 unstack(b19, b13) -> stack(b19, b15)
12 pick-up(b14) -> stack(b14, b13)
13 unstack(b4, b6) -> stack(b4, b1) -> unstack(b6, b17)
    ->stack(b6, b19) -> unstack(b17, b3) -> stack(b17, b14)
14 unstack(b4, b1) -> stack(b4, b6) -> unstack(b1, b2)
    ->stack(b1, b3) -> pick-up(b2) -> stack(b2, b17)
15 unstack(b1, b3) -> stack(b1, b2) -> unstack(b3, b20)
    ->stack(b3, b1) -> pick-up(b20) -> stack(b20, b4)
16 unstack(b3, b1) -> stack(b3, b18) -> unstack(b1, b2)
    ->stack(b1, b20)
17 unstack(b3, b18) -> stack(b3, b2)
18 pick-up(b18) -> stack(b18, b1)
19 pick-up(b9) -> stack(b9, b3)
\end{lstlisting}
\listingcaption{Generated \siwone\ plan for Blocksworld with multiple tower goal and 20 blocks.}
\label{lst:blocks_multiple_2}
\end{figure*}

\begin{figure*}
\footnotesize
\begin{lstlisting}[basicstyle=\footnotesize\ttfamily]
Name: blocks-20-0 (probBLOCKS-20-0.pddl)
\end{lstlisting}
See listing \ref{lst:blocks_multiple_2} for problem specifics.
\begin{lstlisting}
Primitive plan: 74
Plan: 10
1  unstack(b19, b16) -> stack(b19, b18) -> unstack(b12, b13)
    ->stack(b12, b1) -> unstack(b13, b10) -> stack(b13, b19)
    ->unstack(b10, b5) -> stack(b10, b16) -> unstack(b12, b1)
    ->stack(b12, b13) -> unstack(b1, b15) -> stack(b1, b5)
    ->unstack(b15, b14) -> stack(b15, b10)
2  unstack(b1, b5) -> stack(b1, b12) -> unstack(b8, b11)
    ->put-down(b8) -> unstack(b5, b20) -> stack(b5, b8)
3  unstack(b11, b3) -> stack(b11, b5) -> unstack(b3, b17)
    ->stack(b3, b20) -> unstack(b17, b6) -> stack(b17, b3)
    ->unstack(b6, b7) -> stack(b6, b17) -> unstack(b7, b2)
    ->stack(b7, b11) -> unstack(b1, b12) -> stack(b1, b2)
    ->unstack(b6, b17) -> stack(b6, b1) -> unstack(b17, b3)
    ->stack(b17, b14)
4  unstack(b12, b13) -> stack(b12, b7) -> unstack(b13, b19)
    ->stack(b13, b12)
5  unstack(b19, b18) -> stack(b19, b15) -> unstack(b6, b1)
    ->stack(b6, b19)
6  unstack(b18, b4) -> stack(b18, b13) -> unstack(b4, b9)
    ->stack(b4, b6) -> unstack(b3, b20) -> stack(b3, b1)
    ->pick-up(b20) -> stack(b20, b4)
7  unstack(b3, b1) -> stack(b3, b17) -> unstack(b1, b2)
    ->stack(b1, b20) -> unstack(b3, b17) -> stack(b3, b2)
    ->unstack(b18, b13) -> stack(b18, b1)
8  unstack(b17, b14) -> stack(b17, b3) -> pick-up(b14)
    ->stack(b14, b13) -> unstack(b17, b3) -> stack(b17, b14)
9  unstack(b3, b2) -> stack(b3, b18) -> pick-up(b2)
    ->stack(b2, b17) -> unstack(b3, b18) -> stack(b3, b2)
10 pick-up(b9) -> stack(b9, b3)
\end{lstlisting}
\listingcaption{Generated \siwtwo\ plan for Blocksworld with multiple tower goal and 20 blocks.}
\label{lst:blocks_multiple_4}
\end{figure*}

    \clearpage
    \bibliographystyle{ijcai25}
    \bibliography{paper,bib/extra,bib/abbrv,bib/literatur,bib/crossref-short}
\else
\begin{abstract}
In planning and reinforcement learning, the identification of common subgoal structures across problems is important
when goals are to be achieved over long horizons. Recently, it has been   shown that such structures can be expressed
as feature-based rules, called sketches, over a number of classical planning domains.  These sketches
split problems into subproblems which then become solvable in low polynomial time by a greedy sequence of \iw{k} searches.
Methods for learning sketches using feature pools and  min-SAT solvers have
been developed, yet  they face two key limitations: scalability and expressivity.
In this work, we address these limitations by formulating  the problem of learning sketch decompositions
as a deep reinforcement learning (DRL) task,  where  general policies are sought  in  a modified planning
problem  where the  successor states of a state $s$ are defined as those reachable from $s$ through an \iw{k} search.
The sketch decompositions obtained through this method are experimentally evaluated across various domains, 
and problems  are regarded as  solved by the decomposition  when  the goal  is reached  through a greedy sequence of \iw{k} searches.  
While our DRL approach for learning sketch decompositions does not yield interpretable sketches
in the form of rules, we demonstrate that the resulting decompositions can often be understood in a crisp manner.
\end{abstract}

    \section{Introduction}
A common challenge in planning and reinforcement learning is achieving goals that require many actions.
Addressing this challenge typically involves learning useful subgoals or hierarchical policies that abstract primitive actions
\citep{sutton:options,mcgovern-barto-icml2001,josh:hrl,levine:hrl}. Yet the principles underlying
the corresponding  problem decompositions are not well understood. Consequently, methods for learning
subgoals and hierarchical policies often lack robustness, working effectively in some domains
while failing completely in others, without a clear explanation for these differences in
performance.

Recently, a powerful language for expressing, learning, and understanding general problem decompositions has been proposed \citep{bonet:aaai2021,drexler:icaps2022}. A \emph{sketch decomposition} for a class of problems $\Q$ defines a set of subgoal states $G(s)$ for each reachable state $s$ in an instance $P \in \Q$. In any state $s$, the planner's task is not to reach the distant problem goal but to move to a closer subgoal state in $G(s)$.

The concept of a goal being easily reachable or not is formalized through the notion of \emph{problem width} \citep{nir:ecai2012,bonet:jair2024}. A class of problems with width bounded by a constant $k$ can be solved optimally by the \iw{k} algorithm in time exponential in $k$. Many planning domains have a width no greater than 2 when goals are restricted to single atoms. A sketch decomposition $G(\cdot)$ divides problems $P$ in class $\Q$ into subproblems $P[s,G(s)]$, which resemble $P$ but with initial state $s$ and goal states $G(s)$. If all these subproblems have width bounded by $k$, the decomposition width over $\Q$ is bounded by $k$, allowing the problems in $\Q$ to be solved by a greedy sequence of \iw{k} calls, provided the decomposition is acyclic and safe, meaning no subgoal cycles or dead-end states among the subgoals \citep{bonet:aaai2021}.

Methods for learning safe, acyclic sketch decompositions with bounded width, represented by a set of sketch rules, have been developed \citep{drexler:icaps2022,drexler:kr2023}, following techniques previously used for learning general policies \citep{frances:aaai2021}. These learning methods rely on feature pools derived from domain predicates and a min-cost SAT solver, leading to two key limitations: scalability and expressivity. Large feature pools enhance expressivity but result in large theories that are difficult for combinatorial solvers to handle.

In this work, we address these limitations by framing the problem of learning sketch decompositions as one of learning general policies in a deep reinforcement learning (DRL) context\footnote{Data/code available on Zenodo/GitHub:\\
zenodo.org/records/15614893 - github.com/maichmueller/plangolin.}. Here, feature pools are not made explicit, and combinatorial solvers are unnecessary. We build on a novel observation connecting sketch decompositions with general policies and leverage an existing implementation for learning general policies via DRL \citep{simon:kr2023}. The resulting method learns sketch decompositions bounded by a given width parameter $k$ and uses them to search for goals across various domains through a greedy sequence of \iw{k} searches. Unlike symbolic methods, the DRL approach does not produce rule-based sketches but neural network classifiers. However, as we will demonstrate, while interpreting these classifiers is not straightforward, it is often possible to understand the resulting decompositions in a crisp manner.

The structure of the paper is as follows. We begin with an illustrative example and relevant background. We then present the proposed formulation, followed by experiments, an analysis of the decompositions found, related work, and a concluding discussion.

    \section{Example}

The Delivery domain, similar to the Taxi domain in hierarchical reinforcement learning, involves $N$ packages spread across an $n \times m$ grid, 
with an agent tasked to deliver them, one by one, to a target cell. The  sketch decomposition $G_2$,  where $s'  \in G_2(s)$ if the number of
undelivered packages $u(\cdot)$  is smaller in $s'$ than in $s$, yields   subproblems $P[s,G_2(s)]$ of width bounded by $2$, solved optimally by \iw{2}.
The subproblems $P[s,G_2(s)]$ are like $P$ but with initial state $s$ and goal states $G_2(s)$.
Similarly, the sketch decomposition $G_1$,  where $s' \in G_1(s)$ if either $u(s') < u(s)$ and a package is held in $s$, or
$u(s')=u(s)$ and a package is not held in $s$ but is held in $s'$, produces subproblems $P[s,G_1(s)]$ of width 1, solvable optimally by \iw{1}.

\begin{figure}[h]
    \centering
    \begin{minipage}{0.25\textwidth}{
    \hspace{.8cm}\includegraphics[width=0.58\linewidth]{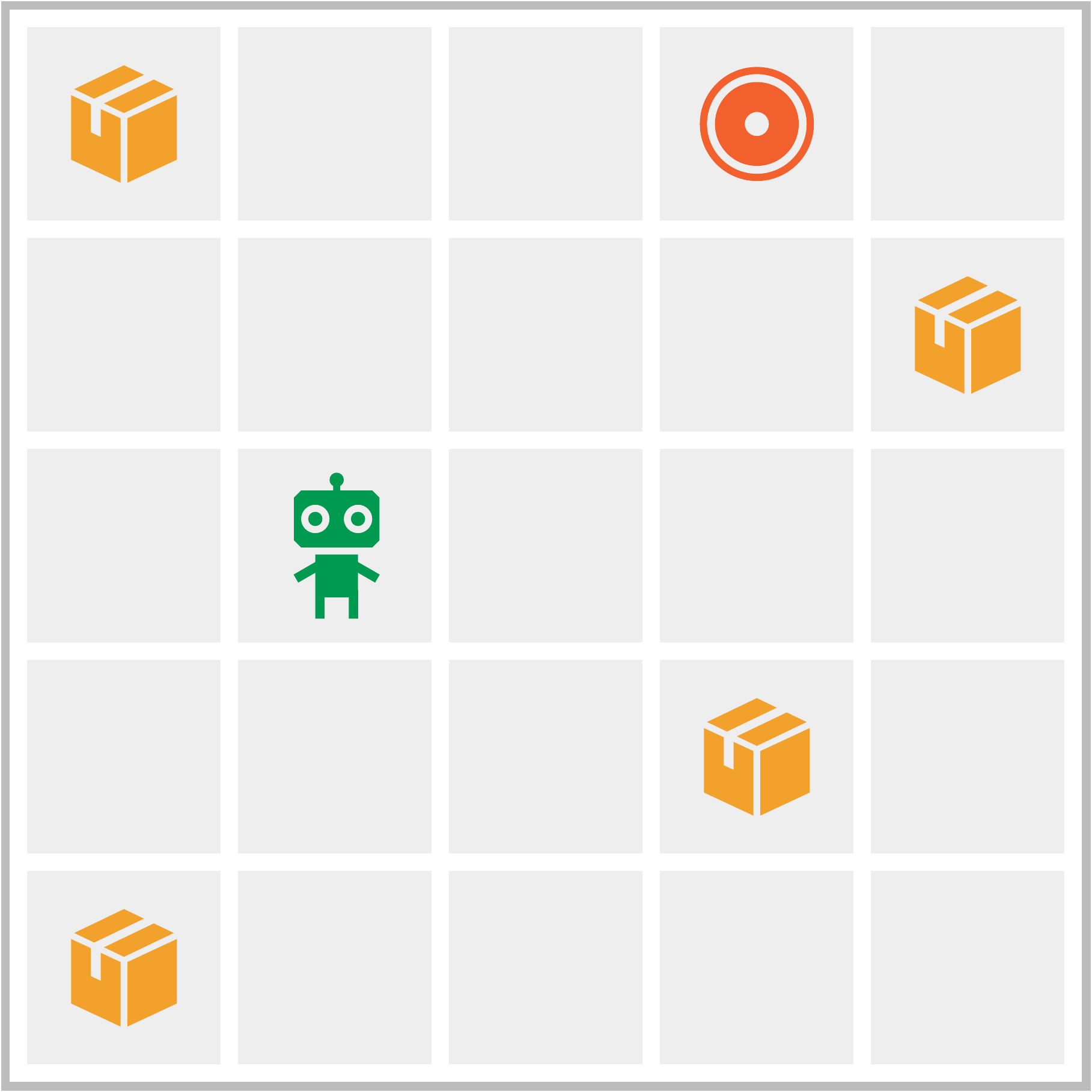}\vspace{0.55cm}
    }\end{minipage}\begin{minipage}{0.4\textwidth}{
    \includegraphics[width=0.5\linewidth,clip, trim={.22cm .3cm 0.5cm .45cm}]{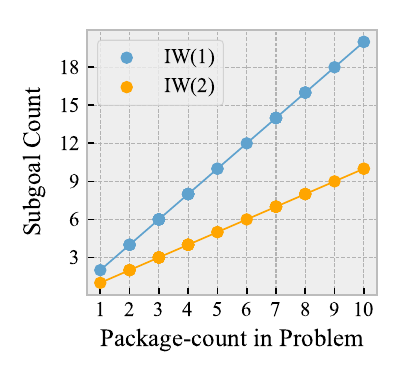}\hfill
    }\end{minipage}
    \caption{\small \emph{Left}. A $5 \times 5$ Delivery instance with $4$ packages, $1$ agent, and target cell (circle). \emph{Right.} Number of subgoals resulting
      from learned decompositions $G_k$  via DRL over test instances as  function of number $N$ of packages. For $k=1$, problems are  solved
      by $2N$  calls to  IW(1); while for $k=2$, by $N$ calls  to  IW(2).}
    \label{fig:example_delivery_decomp}
\end{figure}

A previous  approach learns such decompositions using   feature pools, a width  parameter $k \in \{1, 2\}$, and combinatorial methods
\citep{drexler:icaps2022}. The decompositions are represented implicitly by  collections of \textit{sketch rules}.
This work is aimed  at  learning similar width-k decompositions $G_k(s)$ but using neural networks trained via reinforcement
learning. While the learned representations will not be as transparent, we will see that the resulting decompositions  often are.
Figure \ref{fig:example_delivery_decomp} shows indeed the number of \iw{1} and \iw{2} calls needed to solve Delivery
instances as a function of the number of packages $N$  after learning two  general domain decompositions $G_1$ and $G_2$ via deep reinforcement learning. 
The number of calls, $2N$ and $N$, match exactly the number of calls that would be needed to solve
these  instances using the sketch decompositions defined above, despite using
no rule-based representation or explicit feature pool, but solely a neural net trained via RL.

    \section{Background}

We briefly review  classical planning, the notion of width,  general policies and sketches,
and   methods for learning them, following  \inlinecite{nir:ecai2012}, \inlinecite{frances:aaai2021},
\inlinecite{bonet:aaai2021}, \inlinecite{drexler:icaps2022}, and \inlinecite{simon:kr2023}.

\subsection{Classical and Generalized Planning}

A \emph{planning problem} or \emph{instance} is a pair $P=\tup{\domain,\instance}$
where $\domain$ is a first-order \emph{domain} with action schemas defined over predicates,
and $\instance$ contains  the  objects in the instance and two sets of ground atoms defined
over the objects and predicates defining the initial  and goal situations  $\initial$ and  $\goal$.
An instance $P$ defines a state model $S(P)=\tup{\states,\initialstate,\goalstates,\actions,\applicability,\successor}$ where
the states in $S$ are the possible sets of ground atoms, each one capturing the atoms that are true  in the state.
The initial state $s_0$ is $\initial$, the set of goal states $\goalstates$ are those that include the goal atoms $\goal$,
and the actions $Act$ are the ground actions obtained from the schemas and objects. The ground actions in
$\applicability(s)$ are the ones that are applicable in a state $s$; namely, those whose preconditions are (true) in $s$,
and the state transition function $f$ maps a state $s$ and an action $a \in \applicability(s)$ into the successor state $s'=f(a,s)$.
A \emph{plan} $\pi$ for $P$ is a sequence of actions $a_0,\ldots,a_n$ that is executable in $s_0$ and maps the initial state $s_0$
into a goal state; i.e., $a_i \in \applicability(s_i)$, $s_{i+1}=f(a_i,s_i)$, and $s_{n+1}\in\goalstates$.
A state $s$ is \emph{solvable} if there exists a plan starting at $s$, otherwise it is a \emph{dead-end}. 
The {cost} of a plan is assumed to be given by its length, and a plan is {optimal} if there is no shorter plan.

A  \emph{generalized planning} problem instead is given by a \emph{collection}  $\Q$ of
instances  $P=\tup{\domain,\instance}$ from  a given   domain; for example, all instances
of  Blocks world where the goal just involves $on$ atoms. The solution of a generalized problem
is not an open-loop  action sequence but a closed loop policy as detailed below.
In general, the  instances in $\Q$ are assumed to be solvable, and moreover,
the set $\Q$ is normally assumed to be closed  in the sense that if $P$ is in $\Q$ with initial state $s_0$ and
$P'$ is $P$ but with a solvable initial state reachable from $s_0$, then $P'$ is assumed to be in $\Q$ as well. 

\subsection{Width}

The simplest width-based search procedure is \iw{1}, a modified breadth-first search over the rooted directed graph associated with the state model $S(P)$. It prunes newly generated states that fail to make an atom true for the first time in the search. \iw{k}, for $k > 1$, extends this concept by pruning states that do not make a collection of up to $k$ atoms true for the first time.
These algorithms can be alternatively conceptualized using the notion of state novelty. In this view, \iw{k} prunes states with novelty greater than $k$, where a state's novelty is defined by the size of the smallest set of atoms true in that state and false in all previously generated states.
Central to these algorithms is the concept of problem width. The width of a problem $P$ is determined by the size of the smallest chain of atom tuples $t_0, \ldots t_n$  that is \emph{admissible} in $P$ and has size $\max_i |t_i|$ \citep{nir:ecai2012}. For instance, Blocks World instances with atomic goals $\bon{x}{y}$ and Delivery instances with goals $\sat{pkg}{loc}$ have width $2$ or less.

\iw{k} algorithms find optimal (shortest) solutions in time and space exponential to the problem width. However, planning problems with multiple conjunctive goals often lack a bounded width (\ie, width independent of the instance size). To address this, a variant called \siw was developed. \siw greedily seeks a sequence of \iwraw calls, each decreasing the number $\#g$ of unachieved atomic goals \citep{nir:ecai2012}. It starts with \iw{1}, escalating to IW(2) and beyond if \iw{1} fails to reach a state that decreases $\#g$.
While SIW exploits a particular \emph{problem decomposition} based on unachieved goals, this approach is not universally effective due to potential high-width or unsolvable subproblems. Complete, width-based search algorithms incorporate novelty measures within a best-first search \citep{lipovetzky-et-al-aaai2017,frances-et-al-ijcai2017}.

For convenience, the width of problems $P$ that can be solved in at most one step, is said to have width $0$.
Hence, \iw{0} is defined as breadth-first search that prunes all and only nodes
at depth greater than 1, and \iw{k} is adjusted to never prune nodes at level 1.

\subsection{General Policies and Sketches}

A simple but powerful way to express the solutions to generalized planning problems $\Q$
made up of  a collection of instances $P$, is by means of \emph{rules} of the form $C \mapsto E$
defined over a set of features $\Phi$ \citep{bonet:ijcai2018}. A \emph{state pair} $[s,s']$
satisfies the rule if $C$ is true in $s$ and the features in $\Phi$ change value
when moving from $s$ to $s'$ in agreement with $E$. For example, $E$ can express that
a numerical feature must increase its value, and that a Boolean feature must become
true, etc. A set of rules $R$ defines a non-deterministic \emph{general policy} $\pi$  for $\Q$
which in any reachable state $s$ in $P \in \Q$ selects the successor states $s'$ of $s$
when  the state pair $[s,s']$ satisfies a rule in $R$. The transitions $(s,s')$ are then called \emph{$\pi$-transitions},
and the policy $\pi$ solves an instance $P \in \Q$ if all the $\pi$-trajectories that start in the initial
state of $P$ reach a goal state.
\footnote{These general policies do not map states directly to actions. Instead, they select state transitions and, only indirectly, actions. This choice is convenient for relating general policies and sketches \citep{bonet:aaai2021}.}

The same language used to define general policies can be used to define \emph{sketch decompositions}.
Indeed,  a set of rules $R$ defines the subproblems $P[s,G_R(s)]$ over the reachable non-goal states $s$ of instances $P \in \Q$,
which are like $P$ but with initial state $s$ and goal states $s' \in G_R(s)$ for the 
state pairs  $[s,s']$ that satisfy a rule in $R$. The \emph{width} of the decomposition 
is the maximum width of the subproblems $P[s,G_R(s)]$, $P \in \Q$,  and the decomposition is \emph{safe} and \emph{acyclic} 
in $\Q$ if there is no sequence of (subgoal) states $s_1, \ldots, s_n$,  $s_{i+1} \in G^*_R(s_i)$ and $n > 1$, 
in any $P \in \Q$ that starts in a reachable, alive state $s_1$ (not a dead-end, not a goal),
and ends in a dead-end state $s_n$ or in the same state $s_n=s_1$. Here $G^*_R(s)$ stands for the states $s' \in G_R(s)$
that are closest to $s$. If the decomposition resulting from the rules $R$ is safe, acyclic, and has width bounded by $k$,
then the problems $P \in \Q$ can be solved by a slight variant of the \siw algorithm where a sequence of \iw{k} calls
is used to move iteratively and optimally from a state $s_i$ to a subgoal state $s_{i+1} \in G_R(s_i$), starting from the initial
state of $P$ and ending in a goal state \citep{bonet:aaai2021}.

\subsection{Learning General Policies through DRL}

\begin{algorithm}[t]
  \begin{algorithmic}[1]
    \State \textbf{Input:} Training MDPs $\{M_i\}_i$, each with state priors $p_i$
    \State \textbf{Input:} Policy $\pi(s' \mid s)$ with parameters $\theta$
    \State \textbf{Input:} Value function $V(s)$ with parameters $\omega$
    \State \textbf{Ouput:} Policy $\pi(s' \mid s)$ 
    \State \textbf{Parameters:} Step sizes $\alpha, \beta > 0$, discount factor $\gamma$
    \State Initialize parameters $\theta$ and $\omega$
    \State Loop forever:
    \State\quad Sample MDP index $i \in \{1, \dots, n\}$
    \State\quad\quad Sample non-goal state $S$ in $M_i$ with probability $p_i$
    \State\quad\quad Sample state $S'$ from $N(S)$ with prob. $\pi(S' \mid S)$
    \State\quad\quad Let $\delta = 1 + \gamma V(S') - V(S)$
    \State\quad\quad  $\omega \gets \omega + \beta \delta \nabla V(S)$            
    \State\quad\quad  $\theta \gets \theta - \alpha \delta \nabla \log \pi(S'\mid S)$ 
    \State\quad\quad If $S'$ is a goal state,  $\omega \gets \omega - \beta V(S') \nabla V(S')$
  \end{algorithmic}
  \caption{Actor-Critic RL for generalized planning}
  \label{alg:3:sampled:opt}
\end{algorithm}

Rule-based policies and sketches can be learned without supervision by solving a min-cost SAT problem over the state transitions of a collection of small training instances $P$ from $\Q$ and a pool of features derived from domain predicates and a fixed set of grammar rules based on description logics
\citep{bonet:aaai2019,drexler:icaps2022}. However, some domains require highly expressive features, which necessitate the application of numerous grammar rules, leading to large feature pools that challenge combinatorial solvers.\footnote{A second type of expressive limitation, not discussed here, involves domains where rule-based policies and sketches may require features beyond the capabilities of standard description logic grammars, which are generally fragments of first-order logic with two variables and counting. This limitation also affects non-symbolic approaches based on GNNs \citep{simon:icaps2022,horcik:expressivity}.}

To address this limitation, a recent approach introduced learning general policies in a deep reinforcement learning (DRL) setting \citep{simon:kr2023}. This approach employs a standard actor-critic RL algorithm \citep{sutton:book} with the policy and value functions $\pi(s' \mid s)$ and $V(s)$ represented by neural networks. As shown in Fig.~\ref{alg:3:sampled:opt}, gradient descent updates the parameters $\theta$ and $\omega$ of the policy and value functions. The key difference from standard actor-critic codes is that $\pi$ selects the next state from possible successors $N(s)$ instead of the next action.

The learned policy functions generalize to larger domain instances than those used in training. This generalization is achieved by encoding the policy and value functions in terms of a relational GNN, which generates real-vector embeddings $f^s(o)$ for each object in the instance \citep{gori:gnn,book:gnn}. 
Suitable readout functions then map these embeddings into the values $V(s)$ and probabilities $\pi(s'\mid s)$.

    \section{Learning Decompositions via DRL}

The contribution of this paper is a novel scheme for learning how to decompose planning problems into subproblems that can be solved through iterative applications of the \iw{k} algorithm. Decomposing problems into subproblems is crucial, but the principles guiding such decompositions are not well-defined. Our goal is to achieve decompositions that are general (applicable across a class of problems $\Q$), safe (avoiding dead-ends), acyclic, and have width bounded by $k$. Additionally, we aim to learn these decompositions without relying on combinatorial solvers or explicit feature pools, leveraging the relationship between sketch compositions and general policies, as well as the method reviewed above for learning general policies through DRL.

It is known that general policies are sketch decompositions of zero width, which are safe and acyclic \citep{bonet:jair2024}. Our new observation is that safe, acyclic sketch decompositions with approximate width $k > 0$ for a class of problems $P \in \Q$ can be derived from general policies over a slightly different class of problems $P_k \in \Q_k$, where \emph{the set of successor states $N(s)$ in $P$ is replaced by the set $N_k(s)$ of states reachable from $s$ via \iw{k}}:
\begin{equation}
  N_k(s) \coloneqq \{ s' \mid s'  \text{ is reachable from } s \text{ via \iw{k}} \}.
  \label{nk}
\end{equation}
The \emph{successor states} $s’$ that the policy $\pi(s’ \mid s)$ selects in $P_k$ will be subgoal states $s’$ in $P$ that can be reached from $s$ via \iw{k}. The decomposition’s width is approximately bounded by $k$ because while a width bounded by $k$ guarantees reachability via \iw{k}, the reverse is not necessarily true.\footnote{\citet{nir:ecai2012} refer to reachability via \iw{k} as \textit{effective width} $k$, which is not a robust notion of width, as it is influenced by the order in which child nodes are explored in the breadth-first search.}

The modification of the DRL algorithm from \citep{simon:kr2023} to learn safe and acyclic decompositions of width $k$ over a class of problems $P \in \Q$ is straightforward: the only change required is to replace the set of successor states $N(s)$ in line 10 of Algorithm~\ref{alg:3:sampled:opt} with
the set $N_k(s)$ reachable from $s$ via \iw{k}. This extended set of successor states is then used in the softmax normalization to yield the probabilities $\pi(s’ \mid s)$.
Action costs are assumed to be  all 1  for reaching either  $N$ or $N_k$ successors. 

Let $\pi(s’ \mid s)$ be the general stochastic policy learned by the algorithm in Fig.~\ref{alg:3:sampled:opt} after replacing $N(s)$ with $N_k(s)$.
The resulting decomposition $G_k^\pi(\cdot)$ can then be defined in two ways: \emph{greedily},  as the singleton sets: 
\begin{align}
    G_k^{\pi}(s) &\coloneqq   \{ \, s' \,\} \, , \, s' = \argmax_{s' \in N_k(s)} \pi(s' \mid s) , 
   \label{Gkgreedy}
\intertext{and \emph{stochastically},  as the singleton sets:}
    G_k^{\pi}(s) &\coloneqq   \{ \, s' \,\} \, , \, s' \sim  \pi(s' \mid s), s' \in N_k(s). 
   \label{Gk}
\end{align}
In the first case, a single subgoal state $s'$ for $s$ is chosen as the  \emph{most likely  state} in $N_k(s)$ according to the learned policy $\pi$;
in the second, case, $s'$ is \emph{sampled stochastically} from the set $N_k(s)$ with probability $\pi(s'\mid s)$. 
In $P$, $s’$ may not be a direct successor of $s$ but can be reached from $s$ via \iw{k}. Intuitively,  to  decompose the problem,
we are  allowing  the “agent” to make \iw{k} “jumps” in $P$ following the learned policy for $P_k$ where such ``jumps'' are primitive actions. 
\michael{We need to be clearer here, once we say $\pi$ for the policy we learn then $\pi_k$ for the policy that chooses subgoals (unless I misunderstand?)}

If the policy $\pi$ solves the problem $P_k$, then the decomposition $G_k^\pi$ will be \emph{safe} and \emph{acyclic}. A sequence of subgoal states $s_0, s_1, \ldots, s_n$ with $n > 1$ and $s_{i+1} \in G_k^\pi(s_i)$ for $i = 1, \ldots, n-1$, cannot be cyclic or unsafe, as that would imply the existence of $\pi$-trajectories that do not reach the goal, contradicting the assumption that $\pi$ solves $P_k$. Additionally, this implies the subproblems $P[s, G_k^\pi(s)]$ to have an \emph{approximate} width bounded by $k$, as $G_k^\pi(s) \in N_k(s)$ only includes states reachable from $s$ via \iw{k}.

\medskip

In summary, Algorithm~\ref{alg:3:sampled:opt} is adapted with minor modifications to learn a safe, acyclic, and width-$k$ decomposition $G_k(s)$ for a class of problems $\Q$, though without formal guarantees.\footnote{The symbolic method for learning sketches \citep{drexler:icaps2022} enforces these properties in the training set but cannot guarantee them over the test set. However, this can be addressed manually, case by case.}
The only change involves replacing the set of successor states $N(s)$ in $P \in \Q$ with the set $N_k(s)$ defined in \eqref{nk}. 
The learned stochastic general policy $\pi$ for the resulting class of problems $P_k \in \Q_k$ defines the decomposition $G_k=G_k^\pi$
over $\Q$ as described in \eqref{Gk}.

At test time, the decomposition $G_k^\pi$ is evaluated by running the \iw{k} algorithm sequentially from a state $s$ to a state $s'$ in $G^\pi_k(s)$ until reaching the goal or a maximum number of \iw{k} calls. 
We refer to this algorithm, which applies \iw{k} searches to the decomposition $G^\pi_k(s)$ based on the learned policy $\pi$, as \siwk. 
Unlike the \siw algorithm, \siwk performs a greedy sequence of \iw{k} searches rather than \iwraw searches, requiring each search to end in a state within $G^\pi_k(s)$, not merely a state where the number of unachieved top goals has decreased.
Moreover, unlike \siw, \siwk requires \iw{k} searches to run to completion, as this is necessary for determining the extended set of successors $N_k(s)$ in the decomposition $G^\pi_k(s)$ defined in \eqref{nk}.

    \section{Experiments}

The experiments aim to address several key questions. First, are the learned decompositions $G_k=G_k^\pi$ both general and effective? Specifically, can the (larger) test instances be solved by a greedy sequence of \iw{k} calls? This question is non-trivial, as success with $k=1$ would imply solving instances with linear memory relative to the number of atoms by running \iw{1} sequentially, a significant contrast to solving instances via exponential time and memory search. Second, can the resulting decompositions, represented in the  neural network, be understood and interpreted? This is also challenging, as there is no guarantee that the learned decompositions will be meaningful. To answer the first question, we will examine the coverage of the \siwk algorithm using the learned decomposition $G_k^\pi$, the number of \iw{k} calls (subgoals), and the total plan length. To address the second question, we will analyze plots showing the number of subgoals resulting from the learned decomposition $G_k^\pi$ as a function of relevant parameters of the test instances (\eg, the number of packages). In the experiments, the  subgoal states  $G_k^\pi(s)$ are sampled stochastically
according to \eqref{Gk}, although the results (in the appendix)  are not too different when they are chosen deterministically as in \eqref{Gkgreedy}.
The code and data will be made publicly available.

\paragraph{Learning Setup.}
We use the DRL implementation from \citep{simon:kr2023} with the same hyperparameters to learn the policy $\pi$
that defines the decomposition  $G_k^\pi$. The GNN has feature vectors of size 64 and 30 layers. The Actor-Critic algorithm uses a discount factor $\gamma = 0.999$, a learning rate $\alpha = 2 \times 10^{-4}$, the Adam optimizer \citep{kingma:iclr2015}, and runs on a single NVIDIA A10 GPU for up to 48 hours per domain. Five models are trained independently with different seeds, and the model with the best validation score is selected for testing. The validation score is determined by the ratio $L_V / L_V^*$, where $L_V$ is the plan length from \siwk and $L_V^*$ is the optimal plan length, both averaged over all states of a validation set. Training is stopped early if this ratio approaches 1.0.

\paragraph{Data.}
The domains and training data are primarily from previous works on learning sketches and general policies \citep{drexler:icaps2022,simon:kr2023}. 

This includes Blocks with single and multiple target towers, Childsnack, Delivery, Grid, Gripper, Logistics, Miconic, Reward, Spanner, and Visitall. 
Each domain is tested on 40 larger instances, which extend those used in prior studies
(details in the appendix).

\subsection{Results}
\begin{table*}[]
  \centering
  \footnotesize
  \setlength{\tabcolsep}{2.25pt}

\begin{tabular}{@{\extracolsep{4.8pt}}lrrrrlrrrlr}
\toprule
& \multicolumn{1}{c}{LAMA} & \multicolumn{4}{c}{No Cycle Prevention} & \multicolumn{4}{c}{Subgoal Cycle Prevention} & \multicolumn{1}{c}{Validation} \\ \cmidrule{2-2}\cmidrule{3-6} \cmidrule{7-10} \cmidrule{11-11}
Domain (\#) & Cov. (\%) $\uparrow$ & Cov. (\%) $\uparrow$ & SL $\downarrow$ & L $\downarrow$ & PQ = L / L$_M$  $\downarrow$ & Cov. (\%) $\uparrow$ & SL $\downarrow$& L $\downarrow$ & PQ = L / L$_M \downarrow$ & L$_V$ / L$_V^* \downarrow$ \\
\midrule
&&&&&&& \\
    [-.5em]
& \multicolumn{9}{c}{\textbf{\siwone}} &\\
\midrule
\midrule
Blocks (40) & 40 (\stretchyr{100}{100} \%) & 39 (\stretchyr{100}{98} \%) & 21 & 80 & 1.05 = 80 / 76 & 40 (\stretchyr{100}{100} \%) & 21 & 81 & ~~1.05 = 81 / 77 & 1.22\\
Blocks-mult. (40) & 39 (\stretchyr{100}{98} \%) & 32 (\stretchyr{100}{80} \%) & 19 & 57 & 1.08 = 57 / 53 & 39 (\stretchyr{100}{98} \%) & 23 & 68 & ~~1.15 = 66 / 57 & 1.32\\
Childsnack (40) & 40 (\stretchyr{100}{100} \%) & 40 (\stretchyr{100}{100} \%) & 6 & 11 & 1.06 = 11 / 10 & 40 (\stretchyr{100}{100} \%) & 6 & 11 & ~~1.05 = 11 / 10 & 1.00\\
Delivery (40) & 40 (\stretchyr{100}{100} \%) & 40 (\stretchyr{100}{100} \%) & 10 & 52 & 1.02 = 52 / 50 & 40 (\stretchyr{100}{100} \%) & 10 & 52 & ~~1.02 = 52 / 50 & 1.00\\
Grid (40) & 38 (\stretchyr{100}{95} \%) & 23 (\stretchyr{100}{58} \%) & 7 & 39 & 1.18 = 39 / 33 & 38 (\stretchyr{100}{95} \%) & 71 & 353 & 10.03 = 353 / 35 & 11.85\\
Gripper (40) & 40 (\stretchyr{100}{100} \%) & 40 (\stretchyr{100}{100} \%) & 83 & 165 & 1.33 = 165 / 124 & 40 (\stretchyr{100}{100} \%) & 83 & 164 & ~~1.33 = 164 / 124 & 1.00\\
Logistics (40) & 38 (\stretchyr{100}{95} \%) & 10 (\stretchyr{100}{25} \%) & 8 & 19 & 1.30 = 16 / 12 & 24 (\stretchyr{100}{60} \%) & 113 & 188 & 10.90 = 199 / 18 & 60.36\\
Miconic (40) & 40 (\stretchyr{100}{100} \%) & 40 (\stretchyr{100}{100} \%) & 32 & 60 & 1.15 = 60 / 52 & 40 (\stretchyr{100}{100} \%) & 32 & 61 & ~~1.16 = 61 / 52 & 1.00\\
Reward (40) & 40 (\stretchyr{100}{100} \%) & 40 (\stretchyr{100}{100} \%) & 15 & 197 & 2.32 = 197 / 85 & 40 (\stretchyr{100}{100} \%) & 15 & 196 & ~~2.31 = 196 / 85 & 1.00\\
Spanner (40) & 30 (\stretchyr{100}{75} \%) & 40 (\stretchyr{100}{100} \%) & 24 & 44 & 1.31 = 41 / 31 & 40 (\stretchyr{100}{100} \%) & 24 & 44 & ~~1.30 = 41 / 31 & 1.00\\
Visitall (40) & 40 (\stretchyr{100}{100} \%) & 40 (\stretchyr{100}{100} \%) & 8 & 68 & 1.65 = 68 / 41 & 40 (\stretchyr{100}{100} \%) & 8 & 68 & ~~1.66 = 68 / 41 & 1.03\\

&&&&&&&&&&\\
[-.5em]
& \multicolumn{9}{c}{\textbf{\siwtwo}} &\\
\midrule
\midrule
Blocks (40) & 40 (\stretchyr{100}{100} \%) & 40 (\stretchyr{100}{100} \%) & 9 & 133 & 1.71 = 133 / 77 & 40 (\stretchyr{100}{100} \%) & 9 & 133 & 1.71 = 133 / 77 & 1.27\\
Blocks-mult. (40) & 39 (\stretchyr{100}{98} \%) & 40 (\stretchyr{100}{100} \%) & 8 & 78 & 1.35 = 77 / 58 & 40 (\stretchyr{100}{100} \%) & 8 & 78 & 1.34 = 77 / 58 & 1.07\\
Childsnack (40) & 40 (\stretchyr{100}{100} \%) & 40 (\stretchyr{100}{100} \%) & 3 & 13 & 1.21 = 13 / 10 & 40 (\stretchyr{100}{100} \%) & 3 & 13 & 1.21 = 13 / 10 & 1.00\\
Delivery (40) & 40 (\stretchyr{100}{100} \%) & 40 (\stretchyr{100}{100} \%) & 5 & 57 & 1.12 = 57 / 50 & 40 (\stretchyr{100}{100} \%) & 5 & 56 & 1.12 = 56 / 50 & 1.00\\
Grid (40) & 38 (\stretchyr{100}{95} \%) & 38 (\stretchyr{100}{95} \%) & 3 & 47 & 1.34 = 47 / 35 & 38 (\stretchyr{100}{95} \%) & 3 & 47 & 1.34 = 47 / 35 & 1.00\\
Logistics (40) & 38 (\stretchyr{100}{95} \%) & 40 (\stretchyr{100}{100} \%) & 4 & 34 & 1.33 = 34 / 26 & 40 (\stretchyr{100}{100} \%) & 4 & 34 & 1.33 = 34 / 26 & 1.01\\
    \bottomrule
\end{tabular}
\caption{
\small   Performance metrics for  learned  general decompositions  per domain and width.
  Rows show results over 40 test instances per domain. Coverage (Cov.) indicates the number of successful plans (fraction in parentheses). SL and L represent
  average subgoal and plan lengths  (rounded to nearest integer), while L$^*$ and L$_M$ represent optimal and LAMA plan lengths resp. averaged. 
  Plan quality (PQ) is the ratio of average plan lengths  to those of LAMA.
  Validation score L$_V$ / L$_V^*$ is shown on the right. 
  Arrows indicate preferred metric directions.
}
  \label{tbl:experiments}
\end{table*}

Table~\ref{tbl:experiments} presents the performance of the \siwk algorithm using the learned $G_k^\pi$ decomposition, 
where $\pi$ is the policy derived from the RL algorithm after replacing the set of successors $N(s)$  with $N_k(s)$. 
Key performance metrics include coverage (Cov), subgoal count (SL), and plan length (L). The table’s upper section shows results for \iw{1}, while the lower section displays \iw{2} results for selected domains.

In the table, $L_M$ indicates the plan length computed by the classical planner LAMA, run on an Intel Xeon Platinum 8352M CPU with a 10-minute time and 100 GB memory limit. The columns labeled “subgoal cycle prevention” reflect a minor \siwk algorithm modification that avoids revisiting a subgoal state. For this, 
states that have already been selected as subgoals before are not considered as future subgoals. 
This adjustment impacts performance in  three of the eleven domains, including two  (Grid and Logistics) where the  width-1 decompositions learned
were poor. 

\paragraph{Coverage}

The \siwk algorithm achieves nearly perfect coverage across all domains, except in Logistics, Grid, and Blocksworld-Multiple with \iw{1}. However, for width-2 decompositions, coverage improves to near 100\% in all domains, including these three. By contrast, neither baseline was able to reach this performance.

The reason for this discrepancy in results by width is not entirely clear, but one possibility is that width-1 sketch decompositions cannot be fully captured in the logical fragment represented by GNNs. It is known that GNNs cannot represent width-0 sketch decompositions (\ie, general policies) for Logistics and Grid \citep{simon:icaps2022}, suggesting  that the same might hold for width-1 decompositions. Interestingly, GNNs  do accommodate width-2
sketch decompositions in these domains, as shown in the table, aligning with findings from previous research, which observed that while certain feature pools cannot express rule-based general policies, they can express rule-based sketches.

\paragraph{Subgoal Count}
The column SL in the table presents the average number of subgoals encountered by \siwk on the path to the goal. Although this number alone may not be highly meaningful, it is significantly lower than the average plan lengths, indicating that each subgoal requires multiple actions to be achieved. More interestingly, Figure~\ref{fig:iw1_category1} illustrates the number of subgoals as a function of the number of objects of a selected type per domain (\eg, packages in Delivery, balls in Gripper, children in Childsnack) for $k=1$, and also $k=2$ for Blocks. In the former cases, the relationship is nearly linear, with a coefficient of 1 in Spanner and Reward, and 2 in Delivery, Gripper, Childsnack, and Miconic.
This suggests that the decomposition divides the problem into subproblems, one for each relevant object, which in some cases are further split in two 
(\eg, in Delivery, each package must be picked up and dropped off in separate \iw{1} calls). Despite the use of neural networks and DRL, the resulting
decompositions can be understood. 
In Blocksworld, however, the situation is different. The width-1 decomposition generates more subgoals than there are $\mathit{on}$-atoms in the goal (shown in black versus red), while the width-2 decomposition generates fewer subgoals than
$\mathit{on}$-atoms in the goal (shown in blue). While individual $\mathit{on}$-atoms have a width of 2 and are thus always reachable by IW(2), certain states allow for pairs of $\mathit{on}$-atoms to be reached by \iw{2} as well. The result is that the plots of subgoal counts in Blocksworld are not showing strict linear relationships. 

\paragraph{Plan Quality}
The column $L/L_M$ in the table shows the ratio between the average plan lengths found by \siwk and those found by LAMA. 
Generally, this ratio is close to 1, but there are exceptions in certain domains, such as Reward and Visitall for \iw{1} and several others for \iw{2}.
The general explanation is that decompositions simplify the problems, allowing them to be solved by a linear search like \iw{1}, rather than an exponential search as in LAMA. 
This simplification, however,  can preclude shortcuts, resulting in longer plans. A more specific explanation is that the DRL algorithm minimizes the number of \iw{k} subproblems on the path to the goal, without considering the cost of solving these subproblems, as measured by the number of actions required. 
For instance, if a single package is to be delivered from a state $s$ with two options—a nearby package and a distant one-\siwk does not prefer one over the other since both states $s'$ and $s''$ where either package is held are $N_k$-successors of $s$ with the same cost of 1. Naturally, the \citet{drexler:icaps2022} baseline also exhibits such plan deficits, though to a lesser extent, whereas \citet{simon:kr2023} can surpass LAMA in some domains due to directly optimizing plan cost.

To address this limitation, two approaches could be considered: 1) modifying \siwk to prefer $N_k$-successors $s'$ that have a high probability $\pi(s' \mid s)$ and are closer to $s$, or 2) retaining the true cost information of $N_k$-successors and using this in the DRL algorithm to optimize a combination of the number of subgoals and the cost of achieving them.
The first approach would involve modifying the \siwk algorithm, while the second would require changes to the training process. 
Exploring these approaches, however, is beyond the scope of this work.

It is also worth noting that problem representations (in PDDL) influence problem width and overall plan length. For example, the width-1 decomposition for Gripper results in moving balls from one room to the next one by one, even though two grippers are available. 
This explains why the plans generated by \siwone\ are 33\% longer than those computed by LAMA, which are optimal.
One reason for this inefficiency is that IW(1) cannot load two grippers in a single call; the problem indeed has a width of 2. This situation would differ if the problem representation included a single atom that is true when  both grippers are full. Thus, the ability to learn width-$k$ decompositions is not just limited by the expressive power of GNNs, but also by the problem representation. Making  these limitations and these dimensions  explicit is a strength of our approach, enabling us to understand
the general decompositions that are learned and those that are not or cannot be learned, rather than relying solely on performance metrics.

\paragraph{Baselines}
We also compare our method against two width-$k$ decomposition baselines: the general policy deep RL framework of \citet{simon:kr2023} with $k=0$, and the rule-based combinatorial sketch synthesis approach of \citet{drexler:icaps2022} with $k=1,2$. 

Our method combines aspects of both, as it learns sketch decompositions from deep RL, but not in the form of rules. Empirically, our approach can solve domains in which both baselines fail. For instance, in Logistics and Grid, \siwtwo\ devises near-perfect width-2 decompositions, whereas both baselines fail to achieve substantial coverage. Similarly, in other grid-based environments (Reward, Visitall, Delivery), our decomposition approach attains 100\% coverage, improving over the inconsistent performance of policies by \citet{simon:kr2023}. Overall, our results demonstrate that learning subgoal-based decompositions via DRL imposes less expressive demands than learning general policies and scales better than combinatorial optimization.
Further details on the baselines, can be found in the appendix.  

    \section{Analysis}
\begin{figure*}[t!]
    \centering
    \includegraphics[width=1\linewidth]{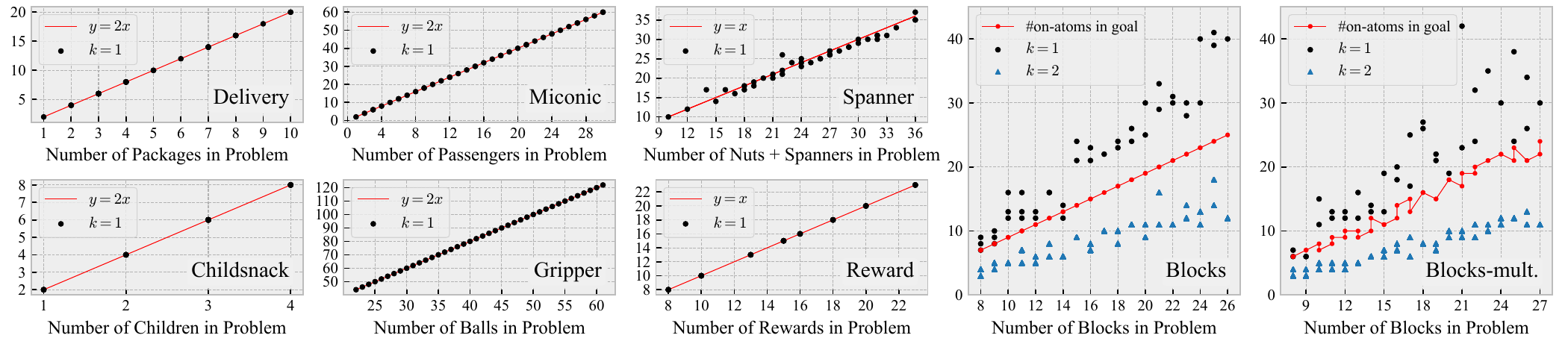}
    \caption{\small %
      Number of subgoals ($y$) generated by \siwk\ as function of characteristic object counts ($x$) in selected domains. 
      The curves are perfect lines in five domains, near perfect in Spanner, and in Blocks \siwtwo\ reaches $\mathit{on}$-atoms in each call, unlike \siwone.
      }
    \label{fig:iw1_category1}
\end{figure*}
The symbolic approach to problem decompositions using sketch rules offers transparency, allowing direct interpretation of the defined decompositions. Interestingly, the inverse process is also possible: equivalent sketch rules can be reconstructed from learned decompositions by analyzing subgoal counts and plan structures. 
Indeed, the width-1 decompositions learned for Delivery, Gripper, Miconic, Childsnack, and Spanner can be understood in terms of four sketches 
with numerical features $N_i$ and Boolean feature $H$:
\begin{align}
\label{eq:ruleset1}
    \{N > 0 \}&\rightarrow \{N\downarrow\} \tag{$R1$}\\
\midrule\midrule
\label{eq:ruleset2}
    \{\neg H , N > 0\} &\rightarrow \{ H \} \tag{$R2$} \\
    \{H , N > 0\} &\rightarrow \{ \neg H, N\downarrow \} \notag \\
\midrule\midrule
\label{eq:ruleset3}
    \{ N_2 >0 \} &\rightarrow \{N_2\downarrow\} \tag{$R3$}\\
    \{N_1 > 0 , N_2 = 0\}   &\rightarrow \{N_1\downarrow\}\notag \\
\midrule\midrule
\label{eq:ruleset4}
    \{N_2 >0 \} &\rightarrow \{N_2\downarrow\} \tag{$R4$}\\
    \{N_1 > 0 , N_2 = 0\}   &\rightarrow \{N_1\downarrow, N_2?\} \notag 
\end{align}
Sketch $R1$ is a simple feature-decrementing sketch, capturing the decomposition in Reward,
where $N$ denotes the number of  uncollected rewards. $R2$ models the decomposition in Delivery, where $H$ indicates whether an undelivered package is held, and $N$ tracks the
number of undelivered packages. $R3$ uses two-counters, and 
prioritizes decrementing $N_2$ until exhaustion, then focuses on $N_1$.
This ruleset characterizes the decomposition of domains like Miconic and Childsnack,
where $N_2$ represents intermediate goals (\eg, sandwiches to make, passengers to board) and $N_1$ denotes final objectives (\eg, children/passengers to serve).

The plans in the Spanner domain follow the sketch $R3$, where the agent first picks up all the spanners (decrementing $N_1$) and then tightens all the nuts (decrementing $N_2$). If this process were perfectly aligned with the sketch, the points in the plot would fall exactly on the line $y = x$. However, the slight deviations from this line allow two observations. Points above the line 
represent unnecessary subgoals where neither $N_1$ nor $N_2$ change.
Points below the line indicate shortcuts in instances with more spanners than nuts where some of the spanners are left uncollected. Often, in these cases however, more spanners are picked up than strictly needed. A possible explanation is that computing the minimum number of spanners to collect, in order to solve the task, may be beyond the expressivity of GNNs given the state encodings.

Finally, $R4$ extends $R3$ with a potential increment of $N_2$ when decrementing $N_1$. These rules describe the decomposition
in Gripper, where $N_1$ represents undelivered balls and N2 denotes balls available for pickup, \ie, the minimum of undelivered balls and free grippers.
Indeed, the model alternates between picking and delivering balls, but only once both grippers hold a ball.

    \section{Related Work}

Methods for decomposing problems into subproblems have been extensively studied in hierarchical planning \citep{htn:planning,nau:shop,aiello:htns}. Hierarchical representations can be derived from precondition relaxations \citep{sacerdoti:precs} and causal graphs \citep{knoblock:hierarchy}, or learned from annotated traces \citep{learning:htn,learning:htn2}. In RL, problem substructure emerges in the form of options \citep{sutton:options}, hierarchies of abstract machines \citep{parr:ham}, MaxQ hierarchies \citep{dietterich:maxq}, reward machines \citep{reward-machines,restraining-bolts}, and intrinsic rewards \citep{singh:intrinsic,singh:intrinsic2}, among others. Although this knowledge is
often provided by hand, methods for learning these structures have leveraged concepts such as "bottleneck states" \citep{mcgovern-barto-icml2001}, eigenvectors of the transition dynamics \citep{machado:eigenpurposes}, and informal width-based notions  \citep{junyent:width}. Additionally, two-level hierarchical policies in DRL have been explored \citep{josh:hrl,levine:hrl}, with the assumption that a master policy can make multi-step "jumps" executed by a low-level worker policy. However, the challenge with bounding these jumps by a fixed number of steps (e.g., 8 steps) is that it fails to capture meaningful substructure that generalizes to larger instances. 

Our approach is closely related to these two-level hierarchies but differs in that the "jumps" are bounded by the concept of width, and instead of being executed by a low-level policy, they are managed by polynomial-time \iw{k} procedures. In principle, these two ideas can be combined. Indeed, symbolic methods for learning hierarchical policies based on width-based considerations have also been developed \citep{drexler:kr2023}.

    \section{Conclusion} 
We have shown that DRL methods can learn subgoal structures for entire collections of planning problems, enabling efficient solutions via greedy IW(k) searches. Though represented by neural networks rather than symbolic rules, these decompositions are often interpretable logically. Our experiments show that decompositions learned from small instances generalize to much larger ones via linear and quadratic \iw{1} and \iw{2} searches. By leveraging width, sketches, and GNN logic, the approach’s limitations can be understood and addressed within this framework.

Two challenges for future work include: (1) incorporating subproblem costs to reduce plan lengths without sacrificing meaningful decompositions, and (2) developing two-level hierarchical policies to avoid \iw{k} searches in subproblems.

    \clearpage
    \ifanonymous
    \else
        \section*{Acknowledgements}
Thanks to Simon Ståhlberg for providing the code of Muninn.
The research has been supported by the Alexander von Humboldt Foundation with funds from the Federal Ministry for Education and Research, Germany.
This project has received funding from the European Research Council (ERC) under the European Union's Horizon 2020 research and innovations programme (Grant agreement No. 885107).
This project was also funded by the German Federal Ministry of Education and Research (BMBF) and the Ministry of Culture and Science of the German State of North Rhine-Westphalia (MKW) under the Excellence Strategy of the Federal Government and the L\"ander.
    \fi
    \bibliographystyle{named}
    \bibliography{paper,bib/extra,bib/abbrv,bib/literatur,bib/crossref-short}
    
    \clearpage

\fi   

\newpage


\end{document}